%% file: main.tex
\documentclass{article}

\usepackage{arxiv}
\usepackage[utf8]{inputenc} 
\usepackage[T1]{fontenc}    
\usepackage{url}            
\usepackage{booktabs}       
\usepackage{amsfonts}       
\usepackage{nicefrac}       
\usepackage{microtype}      
\usepackage{lipsum}
\usepackage{fancyhdr}       
\usepackage{graphicx}       

\usepackage{amsmath}
\usepackage{amssymb}
\usepackage{multirow}
\usepackage{array,colortbl,dcolumn,stfloats}
\usepackage{caption,subcaption}
\usepackage[pagebackref,breaklinks,colorlinks]{hyperref}

\title{RAUM-VO: Rotational Adjusted Unsupervised Monocular Visual Odometry}

\author{
  Claudio Cimarelli, Hriday Bavle, Jose Luis Sanchez-Lopez, Holger Voos \\
  Interdisciplinary Center for Security Reliability and Trust (SnT)\\
  University of Luxembourg  \\
  29, Av. J.F. Kennedy L-1855 Luxembourg\\
  \texttt{name.surname@uni.lu}
}

\usepackage{xspace}
\makeatletter
\DeclareRobustCommand\onedot{\futurelet\@let@token\@onedot}
\def\@onedot{\ifx\@let@token.\else.\null\fi\xspace}

\def\eg{\emph{e.g}\onedot} 
\def\ie{\emph{i.e}\onedot}

\def\etal{\emph{et al}\onedot}
\makeatother

\begin{document}
\maketitle

\begin{abstract}
    Unsupervised learning for monocular camera motion and 3D scene understanding has gained popularity over traditional methods, relying on epipolar geometry or non-linear optimization. Notably, deep learning can overcome many issues of monocular vision, such as perceptual aliasing, low-textured areas, scale-drift, and degenerate motions. Also, concerning supervised learning, we can fully leverage video streams data without the need for depth or motion labels. However, in this work, we note that rotational motion can limit the accuracy of the unsupervised pose networks more than the translational component. Therefore, we present RAUM-VO, an approach based on a model-free epipolar constraint for frame-to-frame motion estimation (F2F) to adjust the rotation during training and online inference. To this end, we match 2D keypoints between consecutive frames using pre-trained deep networks, Superpoint and Superglue, while training a network for depth and pose estimation using an unsupervised training protocol. Then, we adjust the predicted rotation with the motion estimated by F2F using the 2D matches and initializing the solver with the pose network prediction. Ultimately, RAUM-VO shows a considerable accuracy improvement compared to other unsupervised pose networks on the KITTI dataset while reducing the complexity of other hybrid or traditional approaches and achieving comparable state-of-the-art results.
\end{abstract}


\input{latex/content/intro}

\input{latex/content/background}

\input{latex/content/related_works}

\input{latex/content/method}

\input{latex/content/experiments}

\input{latex/content/discussion}

\input{latex/content/conclusion}

\section*{Acknowledgments}
This work was partially funded by the Fonds National de la Recherche of Luxembourg (FNR), under the projects C19/IS/13713801/5G-Sky, and by a partnership between the Interdisciplinary Center for Security Reliability and Trust (SnT) of the University of Luxembourg and LuxConnect S.A.. For the purpose of Open Access, the author has applied a CC BY public copyright license to any Author Accepted Manuscript version arising from this submission.

\bibliographystyle{unsrt}  
\bibliography{egbib}  
\end{document}

%% file: latex/content/intro.tex
\section{Introduction}
\label{sec:intro}

One of the key elements for robot applications is autonomously navigating and planning a trajectory according to surrounding space obstacles. In the context of navigation systems, self-localization and mapping are pivotal components and a wide range of sensors, from exteroceptive as the Global Positioning System (GPS) to proprioceptive as Inertial Measurement Units (IMUs), Light Detection And Ranging (LiDAR) 3D scanners, and cameras, has been employed in the research for a solution to this tasks. As humans experience the rich amount of information coming from vision daily, exploring solutions relying on a pure imaging system is particularly intriguing. Besides, relying only on visual clues is desirable as these are easy to interpret, and cameras are the most common sensor mounted on robots of every kind. 

Visual Simultaneous Localization and Mapping (V-SLAM) methods aim to optimize the tasks of motion estimation, \ie, the 6 Degrees of Freedom (6DoF) transform that relates one camera frame to the subsequent in 3D space, and 3D scene geometry, \ie, depth and structure of the environment, in parallel. Notably, due to the interdependence nature of the two tasks, an improvement on the solution for one influences the other. On the one hand, The mapping objective is to maintain global consistency of the locations of the landmarks, \ie, selected points of the 3D world that SLAM tracks. In turn, revisiting a previously mapped place may trigger a \textit{loop-closure}~\cite{galvez2012bags}, which activates a global optimization step for reducing the pose residual and smoothing all the past trajectory errors~\cite{dellaert2017factor}. On the other hand, Visual Odometry (VO)~\cite{VOTuto} intends to carry out a progressive estimation of the ego-motion without the aspiration of obtaining a globally optimal path. As such, we can define VO as the sub-component of V-SLAM without the global map optimization routine required to minimize drift~\cite{SLAMSurvey}. However, even VO methods construct small local maps composed by the tracked 2D features to which a depth measurement is associated either through triangulation~\cite{PTAM} or probabilistic belief propagation~\cite{VideoBasedSLAM, SemiDenseVO}. In turn, these 3D points are needed to estimate the motion between future frames.

Unsupervised methods have gained popularity for camera motion estimation and 3D geometry understanding in recent years~\cite{DeepVOSurvey}. Especially regarding monocular VO, approaches such as TwoStreamNet~\cite{TwoStreamNet} have shown equally good or even superior performances concerning traditional methods, \eg, VISO2~\cite{VISO2} or ORB-SLAM~\cite{ORB_SLAM}. The unsupervised training protocol~\cite{Zhou17SFMlearner} bears some similarities with the so-called \textit{direct} methods~\cite{DSO}. Both approaches synthesize a time-adjacent frame by projecting pixel intensities using the current depth and pose estimations and minimizing a photometric loss function. However, the learned strategy differs from the traditional because the network incrementally incorporates the knowledge of the 3D structure and the possible range of motions into its weights giving better hypotheses during later training iterations.
Moreover, through learning, we can overcome the typical issues of traditional monocular visual odometry. For example, the support of a large amount of example data during training can help solve degenerate motions (\eg, pure rotational motion), scale ambiguity and scale drift, initialization and model selection, low or homogeneously textured areas, and perceptual aliasing~\cite{SLAMSurvey}. However, aware of the solid theory behind traditional methods~\cite{MVSBook} and their more general applicability, we leverage geometrical image alignment to improve the pose estimation.

Therefore, in this work, we present RAUM-VO. Our approach combines unsupervised pose networks with two-view geometrical motion estimation based on a model-free epipolar constraint to correct the rotations. Unlike recent works ~\cite{TowardsGeneralization, DF-VO} that train optical flow and use complex or computationally demanding strategies for selecting the best motion model, our approach is more general and efficient. First, we extract 2D keypoints using Superpoint~\cite{Superpoint} from each input frame and match the detected features from pairs of consecutive frames with Superglue~\cite{Superglue}. Consequently, we estimate the frame-to-frame motion using the solver proposed by Kneip~\etal~\cite{KneipTwoViewRotBundleAdj}, which we name F2F, and use the rotation to guide the training with an additional self-supervised loss. Finally, RAUM-VO efficiently adjusts the rotation predictions with F2F during online inference while retaining the scaled translation vectors from the pose network.

\begin{figure*}[htpb]
        \centering
		\includegraphics[width=\columnwidth]{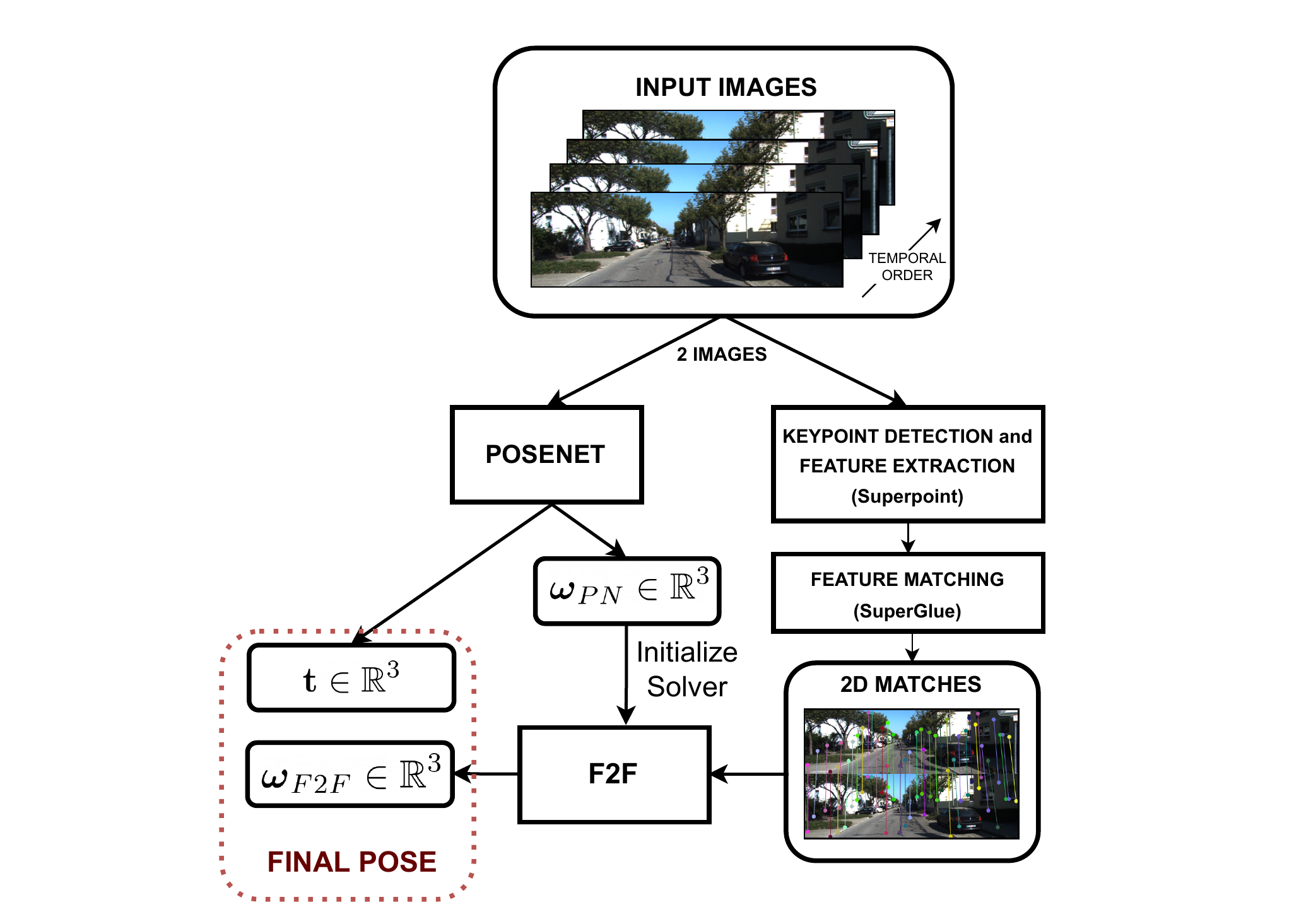}
		\caption{\textbf{RAUM-VO block diagram}. The figure shows the flow of information inside RAUM-VO from the input image sequence to the final estimated pose between each pair of consecutive image frames.}
		\label{fig:diagram}
\end{figure*}

Ours contributions are summarized as follows:

\begin{itemize}

\item We present RAUM-VO, an algorithm to improve the pose estimates of unsupervised pose networks for monocular odometry. To this end, we introduce an additional self-supervision loss using frame-to-frame rotation to guide the network's training. Further, we adjust the rotation predicted by the pose network using the motion estimated by F2F during online inference to improve the final odometry.


\item We compare our method with state-of-the-art approaches on the widely adopted KITTI benchmark. RAUM-VO improves the performance of pose networks and is comparably good as more complex hybrid methods while being more straightforward to implement and more efficient.


\end{itemize}

%% file: latex/content/background.tex
\section{Background on SLAM}

The difference between SLAM and VO is the absence of a mapping module that performs relocalization and global optimization of the past poses. Aside from this aspect, we can consider contributions in monocular SLAM works seamlessly with those in VO literature. A primary type of approach to SLAM is filter-based, either using Extended Kalman Filters (EKFs) as in MonoSLAM~\cite{Monodepth2} or Particle Filters as in FastSLAM~\cite{FastSLAM} and keyframe-based~\cite{PTAM}, referred in robotics to as \textit{smoothing}~\cite{SqrootSAM}. This name entails the main difference between keyframe-based and filtering. While the first optimizes the poses and the landmarks associated with keyframes (a sparse subset of the complete history of frames) using batch non-linear least squares or Bundle Adjustment (BA)~\cite{BundleAdj}, the latter marginalizes past poses' states to estimate the last at the cost of accumulating linearization errors~\cite{VIOAR}. In favor of bundle adjustment, Strasdat~\etal~\cite{WhyFilter} show that the accuracy of the pose increases when the SLAM system tracks more features and that the computational cost for filtering is cubic in the number of features' observations compared to linear for BA. Thus, using BA with an accurate selection of keyframes allows more efficient and robust implementations of SLAM. Unsupervised methods are more similar to the keyframe-based SLAM. The motion is not the result of a probabilistic model propagation and a single-step update but of an iterative optimization to align a batch of image measurements.

Motion estimation approaches fall either into \textit{direct} or \textit{indirect} categorization based on the information or measurements included in the optimized error function. The direct method~\cite{LSD-SLAM, DSO} includes intensity values in a non-linear energy function representing the photometric difference between pixels or patches correspondences. These are found projecting points from one frame to another using the current motion and depth estimation, which is optimized either through Gauss-Newton or Levenberg-Marquardt. Instead, indirect methods~\cite{PTAM, ORB_SLAM} leverage epipolar geometry theory~\cite{MVSBook} to estimate motion from at least five matched 2D point correspondences in case of calibrated cameras~\cite{Nister5points} or eight for the uncalibrated case~\cite{longuet1981computer}. After initializing a local map from triangulated points, Perspective-n-Point (PnP)~\cite{EPNP} can be used with \textit{Random Sample Consensus} (RANSAC) robust iterative fitting scheme~\cite{cantzler1981random} to obtain a more precise relative pose estimation. Subsequently, local BA refines the motion and the geometrical 3D structure by optimizing the reprojection error of the tracked features.

We do not apply the BA technique to correct the accumulated pose errors in this work. However, we investigate PnP motion estimation in place of the trained pose network and compare the results in Section~\ref{sec:gc}.

%% file: latex/content/related_works.tex
\section{Related Work}
\label{sec:rel_w}

\subsection*{Unsupervised Learning of Monocular VO}

The pioneering work of Garg~\etal~\cite{Garg16} represents a fundamental advancement because they approached the problem of depth prediction from a single frame in an unsupervised manner for the first time. Their procedure consists of synthesizing a camera's depths in a rectified stereo pair by warping the other using the calibrated baseline and focal lengths. Godard~\etal~\cite{Godard17} use the stereo pair to enforce a consistency term between left and right synthesized disparities while adopting the \textit{Structural Similarity} (SSIM) metric~\cite{SSIM} as a more informative visual similarity function than the $L_1$ loss.
SfM-Learner~\cite{Zhou17SFMlearner} entirely relies on monocular video sequences and proposes to use the bilinear differentiable sampler from ST-Nets~\cite{SpatialTransformerNet} to generate the synthesized views.

Because the absolute metric scale is not directly observable from a single camera (without any prior about object dimensions), stereo image pairs are also helpful to recover a correct metric scale during training while maintaining the fundamental nature of a monocular method~\cite{UndeepVO, UnsupWFeatRec, Monodepth2}. Mahjourian~\etal~\cite{UnsupMono3dICP} imposes the scale consistency between adjacent frames as a requirement for the depth estimates by aligning the 3D point clouds using \textit{Iterative Closest Point} (ICP) and approximating the gradients of the predicted 6DoF transform. Instead, Bian~\etal~\cite{Bian19}, arguing that the previous approach ignores second-order effects, show that it is possible to train a globally consistent scale with a simple constraint over consecutive depth maps, allowing to reduce drift over long video sequences. In~\cite{Luo20Consistent}, a \textit{Structure-from-Motion} (SfM) model is created before training and used to infer a global scale while the image space distance between projected coordinates and optical flow displacements. More recently, several approaches~\cite{TowardsGeneralization, DF-VO, GeneralizingDeepVO} have leveraged learned optical flow dense pixel correspondences to recover up-to-scale two-view motion based on epipolar geometry. Therefore, they resolve the scale factor by aligning a sparse set of points with the estimated depths.

One of the main assumptions of the original unsupervised training formulation is that the world is static. Hence, many works investigate informing the learning process about moving objects through optical flow~\cite{casser2019depth, vijayanarasimhan2017sfm, yin2018geonet, zou2018dfnet, zhao2018learning, lee2019learning, ranjan2019competitive, EPC++, chen2019self, li2020unsupervised, wang2021motionhint, jiang2021unsupervised}. The optical flow, which represents dense maps of the pixel coordinates displacement, can be separated into two components. The first, the rigid flow, is caused by the camera's motion. The second, the residual flow, is caused by dynamic objects that move freely to the camera frame. Therefore, these methods train specific networks to explain the pixel shifts inconsistent with the two-view rigid motion. However, these methods focus principally on the depth and optical flow maps quality and give few details about the impact of detecting moving objects on the predicted two-view motion. Notably, they use a single metric to benchmark the relative pose that is barely informative about the global performance and cannot distinguish the improvements clearly.

A recent trend is to translate traditional and successful approaches such as SVO~\cite{SVO}, LSD-SLAM~\cite{LSD-SLAM}, ORB-SLAM~\cite{ORB_SLAM} and DSO~\cite{DSO} into their learned variants or to take inspiration for creating hybrid approaches where the neural networks usually serve as an initialization point for filtering or Pose Graph Optimization (PGO)~\cite{wang2018learning, yang2018deep, li2019pose, loo2019cnn, tiwari2020pseudo, cheng2020depth, bian2021unsupervised, D3VO}. However, RAUM-VO focuses on improving the predicted two-view motion of the pose network without introducing excessive computations overhead as required by a PGO backend.


 Instead of training expensive optical flow, RAUM-VO leverages pre-trained Superpoint~\cite{Superpoint} network for keypoint detection and feature description and Superglue~\cite{Superglue} for finding valid correspondences. Unlike optical flow, the learned features do not depend on the training dataset and generalize to a broader set of scenarios. Also, using Superglue, we avoid heuristics for selecting good correspondences among the dense optical flow maps, which we claim could be a more robust strategy. However, we do not use any information about moving objects to discard keypoints lying inside these dynamic areas. Finally, differently from other hybrid approaches~\cite{TowardsGeneralization, DF-VO}, we do not entirely discard the pose network output, but we look for a solution that improves its predictions efficiently and sensibly. Thus, the adoption of the model-free epipolar constraint of Kneip and Lynen~\cite{KneipTwoViewRotBundleAdj} allows us to find the best rotation that explains the whole set of input matches without resorting to various motion models and RANSAC schemes.
 To the best of our knowledge, we are the first to test such an approach combined with unsupervised monocular visual odometry.

%% file: latex/content/method.tex
\section{Method}
\label{sec:meth}

\begin{figure*}[htpb]
        \centering
		\includegraphics[width=0.9\textwidth]{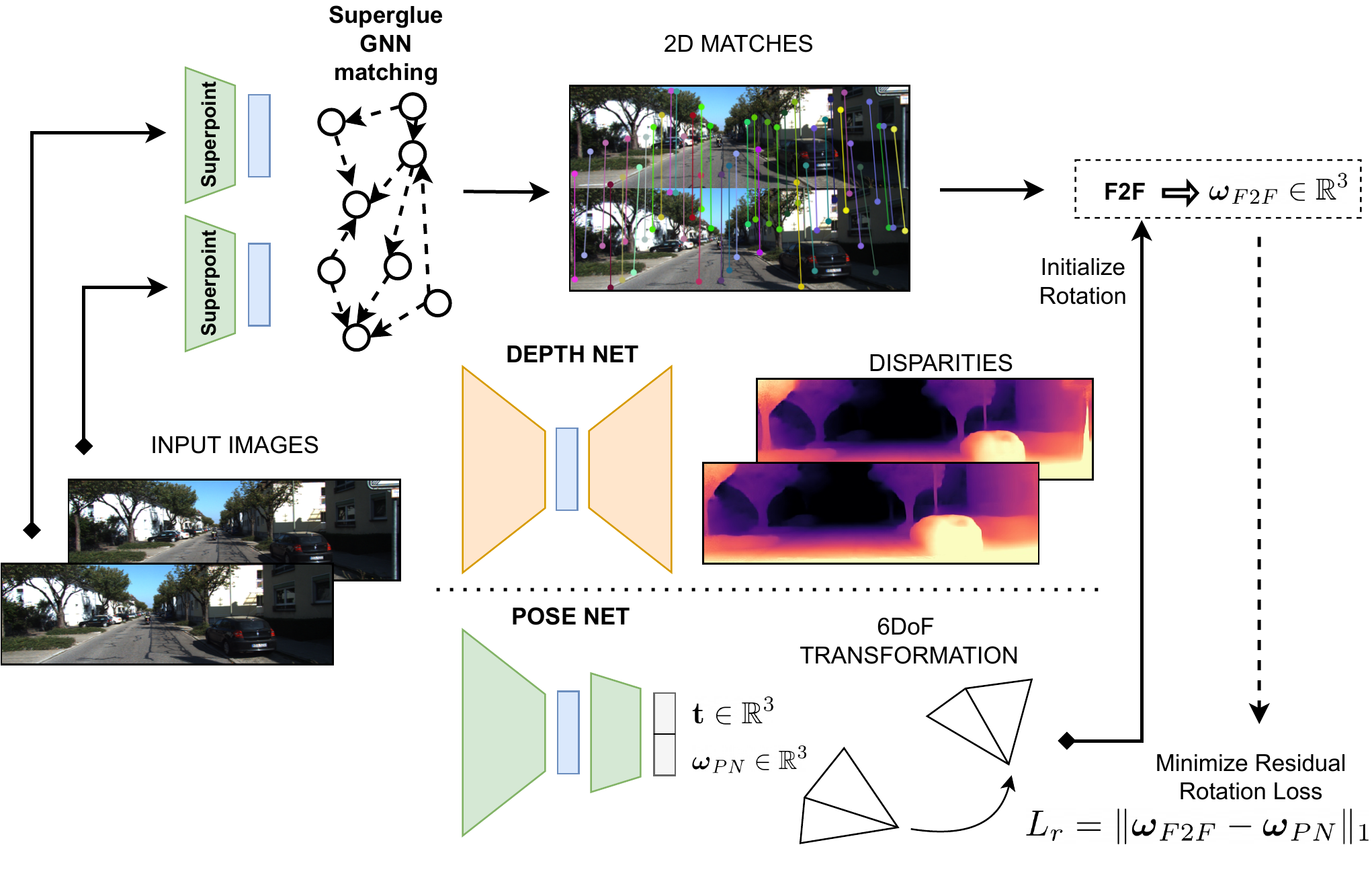}
		\caption{\textbf{Diagram of RAUM-VO training}. A sequence of images and 2D matches between pairs is the input for the training. The Depth Network takes only a single image to output a disparity map. The Pose Network outputs the 3D rigid transformation, as rotation and translation, between the two input images temporally ordered concatenated along the channel dimension. The matches are the input to the frame to frame rotation algorithm, whose output guides the training and adjusts the Pose Network estimation at test time.}
		\label{fig:pipeline}
\end{figure*}

This section outlines the proposed algorithm, RAUM-VO, for estimating the motion from a sequence of monocular camera images using a combination of deep neural networks and traditional epipolar geometry. This work follows Zhou~\etal~\cite{Zhou17SFMlearner} established unsupervised training protocol based on view synthesis and photometric loss, which we describe in Section~\ref{sec:viewsy}. In addition, to facilitate the learning process, we describe additional techniques implemented in our training in Sections~\ref{sec:dm} and~\ref{sec:dc}. As shown in Figure~\ref{fig:pipeline}, the training outcome is a Depth Network that has learned to associate a disparity map to a single input image frame and a Pose Network that predicts the 6DoF rigid transformation between two consecutive frames. Additionally, we use the Superpoint~\cite{Superpoint} network to extract 2D keypoints descriptors. Consequently, using a pre-trained Superglue Graph Neural Network (GNN)~\cite{Superglue}, RAUM-VO matches the corresponding features between pairs of successive frames. These matches are the input for the two-view motion estimation method~\cite{KneipTwoViewRotBundleAdj} (see Section~\ref{sec:f2f}), whose rotation corrects the network's output. 

\subsection{View Synthesis and Photometric Loss}
\label{sec:viewsy}

The principle for obtaining a supervision signal shares some similarities with direct visual odometry~\cite{wang2018learning}. Given to images at time $t$ and $t+1$, $I_t$ and $I_{t+1}$ respectively, the Depth Network produces a disparity (inverse depth) maps $d_t$, resp. $d_{t+1}$, and the Pose Network a 6DoF transformation $\mathbf{T}_{t\rightarrow t+1} = [\ \mathbf{R}\ |\ \mathbf{t}\ ]$. Then, we obtain the depth maps $D_t$ and $D_{t+1}$ by inverting the disparities and normalizing them between a predefined minimum and maximum range limit. Finally, let $\mathbf{K}$ denote the intrinsic camera matrix, and $\mathbf{p}_t = [u, v] $ a 2D pixel coordinate on $I_t$ image plane. in 2D homogeneous coordinates. The projection of $\mathbf{p}_t$ into the reference frame of $I_{t+1}$, $\mathbf{p}_{t\rightarrow t+1}$, is given by the following equation:

\begin{equation}
 \mathbf{p}_{t\rightarrow t+1} = \pi( \mathbf{K} \mathbf{T}_{t\rightarrow t+1} \mathbf{K}^{-1} \mathcal{H}(\mathbf{p}_t, D_t[\mathbf{p}_t]))\,,
\end{equation}

where $D_t[\mathbf{p}_t]$ denotes the depth value at the point $\mathbf{p}_t$. $\mathcal{H}$ the operation to lift the 2D pixel coordinates to 3D homogeneous coordinates:
\begin{equation}
   \mathcal{H}\colon  ([u, v], z)\mapsto [u*z, v*z, z, 1] = [x, y, z, 1]\,,
\end{equation}

and $\pi$ is the projection to the image plane:

\begin{equation}
     \pi\colon ([x, y, z, 1]) \mapsto  [x/z, y/z] = [u,v]\,.
\end{equation}

Using the (sub-)differentiable bilinear sampling operation, that we note with $\mathcal{S}$, introduced with Spatial Transormer Networks (STNs)~\cite{SpatialTransformerNet}, we obtain a synthesized version of $I_{t+1}$, $I_{t\rightarrow t+1}$, by interpolating its intensity values at the locations indicated by a grid of points $\mathbf{p}_{t\rightarrow t+1}$.
\begin{equation}
 I_{t\rightarrow t+1} = \mathcal{S}(I_{t+1}, \mathbf{p}_{t\rightarrow t+1})\,.
\end{equation}

Next, we optimize the estimated disparities and poses by minimizing the perceptual distance between the image $I_{t+1}$ and its synthesized version $I_{t\rightarrow t+1}$. Following the initial suggestion of~\cite{visualloss} and the example of previous similar works~\cite{UndeepVO, Godard17}, this distance is best assessed by a combination of L1 and SSIM~\cite{SSIM}, which is differentiable with respect to both Depth and Pose Networks parameters. Particularly, the SSIM function aims to quantify the visual similarity of $I_{t+1}$ and its synthetic reconstruction $I_{t\rightarrow t+1}$ by comparing the luminance, contrast, and structure measurements on windows of size $n\times n$. 

Therefore, the \textit{photometric loss}, $L_p$, equates to:
\begin{equation}
    \begin{split}
 L_p =&\; \alpha_\mathrm{SSIM} \frac{1-\mathrm{SSIM}(I_{t+1}, I_{t\rightarrow t+1})}{2}\; +\\
 &\; \alpha_{l_1}\|I_{t+1}-I_{t\rightarrow t+1}\|_1\,.
    \end{split}
\end{equation}
In our experiments, we set $\alpha_\mathrm{SSIM} = 0.85$ and $\alpha_{l_1} = 0.15$.

Notably, this warping mechanism succeeds with the assumption that the scene is static, there are no occlusions, and the lighting conditions are constant without reflections. Notwithstanding the training process may be robust to minor violations of these assumptions, solutions for reducing dynamic objects~\cite{EPC++} and non-Lambertian surfaces~\cite{D3VO} impact on the optimization convergence have been provided in the recent literature. 
Instead, we rely on simpler mechanisms to alleviate the dynamic world conditions. During training, we extend the view synthesis procedure to the previous frame $I_{t-1}$ as well. Hence, we consider the minimum between $L_p(I_{t-1},I_{t})$ and $L_p(I_{t-1},I_{t})$ on a per-pixel basis as the final photometric loss. This strategy mitigates the effects of dis-occluded pixels~\cite{Monodepth2}.

To conclude, we would like to add a few observations. First, while the output would be random at the beginning, it is expected to converge to a meaningful value through the joint optimization process of the two networks. Next, the scale of the 6DoF transformation, foreseeably, reflects the depth scale as they are jointly optimized. But, even if not aligned with the metric scale of the scene, it is plausibly globally consistent. Remarkably, this is an advantage over geometrical methods since, for the latter, we would need to take further precautions to avoid scale drifts~\cite{strasdat2010scale,LSD-SLAM}. In Section~\ref{sec:dc}, we will introduce an additional loss term to reinforce a global consistency constraint during training.

\subsection{Depth Smoothness Loss}
\label{sec:dm}

The photometric loss is not informative with homogeneous or low-textured areas of an image, and the depth estimation problem becomes ill-posed. The pixels in these regions can be associated with disparity values and still obtain a similar visual appearance for a fixed rigid transformation~\cite{Monodepth2}. However, we can introduce a prior on the estimated depth maps that encourage smooth changes of the disparities inside these regions while discouraging the formation of holes. Thus, by considering the first (or second~\cite{wang2018learning}) order gradients of the image as weighting terms, we allow sharp discontinuities to appear only in correspondence of edges~\cite{Godard17}. 

Therefore, the following equation constitutes the \textit{smoothing loss} $L_s$:
\begin{equation}
    L_s = \left | \partial_x d_t   \right | e^{-\left | \partial_x I_t \right |} + \left | \partial_y d_t   \right | e^{-\left | \partial_y I_t \right |}\, ,
\end{equation}

where $\partial_x$ and $\partial_y$ are the first derivatives of the color image and disparity map taken along $x$ and $y$ directions.

\subsection{Depth Consistency Loss}
\label{sec:dc}

An issue of monocular VO, famously, is the non-observability of the metric scale of the surrounding environment and, consequently, of the motion between two views. This limitation leads to the well-known issue of scale-drift, which has been successfully addressed in traditional BA-SLAM by performing the pose graph optimization over 3D similarity transforms~\cite{strasdat2010scale,LSD-SLAM}. From the perspective of learned mono-VO, Tateno~\etal~\cite{CNN-SLAM} explore the path of predicting depth maps using CNNs, confident of their capability to reproduce the metric scale passed through the ground-truth depths supervision. On the other hand, without depth supervision, an alternative approach to learning a metrically scale-aware network is from information regarding the translation vectors norm, as in~\cite{3dpacking} where the authors impose a velocity loss. Even though we cannot obtain the real scale during training, ensuring the depth consistency is fundamental for reducing the drift and easing the task of aligning the estimated trajectory with an external metric map. Therefore, in this work, lacking the knowledge of real-world scale and ground-truth depths, we adopt the loss for imposing depth consistency between two frames introduced by Bian~\etal~\cite{Bian19}. The following equation defines the \textit{depth consistency loss} $L_{dc}$:

\begin{equation}
    L_{dc} = \frac{\left |D_{a\rightarrow b} - D_b \right |}{D_{a\rightarrow b} + D_b }\, ,
\end{equation}
where $D_{a\rightarrow b}$ represent the synthesized version of the depth estimated for image $I_a$ to the camera reference of image $I_b$ by means of the estimated pose $\mathbf{T}_{a\rightarrow b}$ and the bilinear sampler.

\subsection{F2F: Frame to Frame Motion}

\label{sec:f2f}

Herewith, we describe the pivotal component of our proposed method. In particular, we incorporate the rotation optimization formulated by Kneip and Lynen~\cite{KneipTwoViewRotBundleAdj}. They propose an alternative epipolar constraint that enables to solve the relative pose problem without many of the issues encountered in Essential-matrix based methods. Namely, these are:
\begin{itemize}
\item the indirect parametrization of the motion that has to be decomposed from the essential matrix as~\cite{MVSBook}:
 \begin{equation}
     \mathbf{E} = [\mathbf{t}]_x\ \mathbf{R}\,;
 \end{equation}
 \item multiple solution from the decomposition that have to be disambiguated through a cheirality check and hence by triangulation;
 \item degenerate solutions that may result from either points lying on a single planar surface, distribution of the points in a small image area, and pure translational or rotational motion. In these cases, one approach is to select a different motion model,~\eg the Homography matrix, after identifying the degeneracy with a proper strategy.
\end{itemize}

So, given a set of image points $(\mathbf{p}_i, \mathbf{p}_i')$ matched between two views, we translated them into pairs of unit bearing vectors $(\mathbf{f}_i, \mathbf{f}_i')$ through normalization. These vectors ideally start from the camera center and point in direction of the corresponding 3D points and each pair defines an epipolar plane. Then, the authors observe that the all the normal vectors of the epipolar planes need to be coplanar~\cite{TwoViewRotation}. The normals vector forms together a 3-by-$n$ matrix $\mathbf{N}=[\mathbf{n}_1\ \ldots \ \mathbf{n}_n]$, and are defined as follows:
\begin{equation}
    \mathbf{n_i} = \mathbf{f_i} \times \mathbf{R}\mathbf{f_i'}\,.
\end{equation}
Due to the coplanarity constraint, the covariance matrix $\mathbf{NN}^T =\ \mathbf{M}$ has to be at most of rank 2. Notably, the problem is equivalent to a rank minimization parametrized by $\mathbf{R}$, and is solved by finding the matrix $\mathbf{M}$ with the smallest minimum \textit{eigenvalue}:
\begin{equation}
    \mathbf{R}\ = \ \underset{\mathbf{R}}{\mathrm{\arg\min}} \lambda_{\mathbf{M}, \mathrm{\min}}\,.
\end{equation}
Furthermore, the authors observe that the \textit{eigenvector} associated with $\lambda_{\mathbf{M}, \mathrm{\min}}$ corresponds to the translation direction vector. Therefore, this method, which we name \textbf{F2F}, is able to retrieve the full frame-to-frame motion.

The problem is solved with a Levenberg-Marquardt procedure. To avoid the possible presence of local minima typical of non-linear optimization, we use the rotation estimated by the Pose Network as starting point. In Section~\ref{sec:gc}, we show the benefits of this initialization. Also, we choose to perform a single optimization with all the matches instead of multiple RANSAC iterations. For restricting the number of matches outliers, we set the threshold of the Superglue match confidence score to $0.9$. At the moment, we found that this approach works best for the data into hands after empirical evaluation of multiple RANSAC settings and inlier criteria. 

Lastly, we include the rotation $\mathbf{R_{F2F}}$ as supervision for the rotation output of the Pose Network, $\mathbf{R_{PN}}$, in \textit{residual rotation loss} $l_{r}$. To this aim,  we map the rotation matrices into their axis-angle counterparts through the logarithm function:
\begin{equation}
        \textstyle{\log} \colon \mathrm {SO} (3) \to \textstyle{\mathfrak {so}}(3);\; \mathbf{R} \mapsto \textstyle{\log}(\mathbf{R})\,,
\end{equation}

where $\textstyle{\mathfrak {so}}(3)$ is the Lie algebra associated to the Lie Group of 3D rotations $SO(3)$~\cite{gao2021introduction}. Based on the isomorphism between ${\mathfrak {so}}(3)$ and $\mathbb{R}^3$ with the cross product, we treat the logarithm of a rotation matrix as a vector $\omega \in \mathbb{R}^3 $ decomposed into a unit-norm direction vector $\mathbf{u}\in \mathbb{R}^3 $, representing the rotation axis, and its $L_2$ norm $\theta\in \mathbb{R}$, where $\theta \in [0, \pi]$ represents the angle of rotation:

\begin{equation}
    \textstyle{\log}(\mathbf{R}) = \textstyle {\boldsymbol {\omega }} = \theta \mathbf{u} \,. 
\end{equation}

Therefore, we can compute the $L_1$ norm, denoted by $\lVert \cdot \rVert_1$, of the distance between the rotation vector predicted by the network, ${\boldsymbol {\omega }}_{PN} $, and the one estimated by F2F, ${\boldsymbol {\omega }}_{F2F} $. Thus, we obtain the following rotation loss, $L_r$:

\begin{equation}
    L_r  = \| \textstyle {\boldsymbol {\omega }}_{F2F} - \textstyle {\boldsymbol {\omega }}_{PN} \|_{1} \,.
\end{equation}

In Figure~\ref{fig:pipeline}, we show how all the components we described interact during the training of RAUM-VO.

The implementation of F2F used in this work is the one provided by the OpenGV library~\cite{kneip2014opengv}.

%% file: latex/content/experiments.tex
\section{Experiments}

\label{sec:exp}

This Section provides details regarding our experimental procedure and the settings for accurately reproducing our results. Also, we provide the results of VO obtained on KITTI and compare with state-of-the-art methods.


\subsection{Training Procedure}
\label{sec:training}

Because we have experienced a degrade in performance when including the $l_{dc}$ term early in training, we split it into two phases. Particularly, when the Depth Network has not yet found a convergence direction for a plausible geometrical structure, the $l_{dc}$ term, especially if with a magnitude outweighing the photometric loss norm, could cause the depth maps to collapse towards a local minimum during the initial training phase. An alternative solution may be to adjust adaptively the weighing of $l_{dc}$ based on the value of $l_{p}$. Therefore, we add the depth consistency loss after the convergence of the photometric loss. Also, we add the contribution of the loss $l_{r}$ in the second training phase to let the Pose Network reach an initial convergence plateau first. 

Consequently, we obtain two models:
\begin{itemize}
    \item \textbf{Simple-Mono-VO} is obtained after the first training phase by selecting the checkpoint with the best $t_{err}$ on the training set;
    \item \textbf{RAUM-VO} is obtained after the second phase by selecting the checkpoint with the best $t_{err}$ on the training set and correcting the rotations with the output of F2F.
\end{itemize}

\subsection{Networks architectures}

The \textbf{Depth Network} has an encoder-decoder architecture~\cite{UNET} with skip connection similar to the \textit{DispNet}~\cite{mayer2016large} used by SfM-Learner~\cite{Zhou17SFMlearner}. Specifically, the encoder is a ResNet18~\cite{resnet}, and the decoder has five layers of 3$\times$3 convolutions followed by an ELU activation function~\cite{elu}, an up-sampling, and a concatenation with the ``connected'' encoder feature. In accordance with~\cite{Bian19}, we avoid the multi-scale training for efficiency purposes. Therefore, we apply the Sigmoid function to the last output to obtain a disparity map.

The \textbf{Pose Network} consists of one ResNet18~\cite{resnet} encoder that takes as input an pair of images concatenated along the channel dimension. The feature extracted by the last layer is then the input to a small CNN decoder composed by:
\begin{enumerate}
    \item one Linear layer that reduces the feature to a 256-dim vector followed by ReLU~\cite{relu} non-linearity;
    \item two Convolutional layers with 256 kernels of size 3$\times$3 followed by ReLu non-linearities;
    \item one Linear layer that outputs the 6DoF pose vector as the vector $\mathbf{x}\in \mathbb{R}^6$, which contains the concatenation of the translation $\mathbf{t}\in \mathbb{R}^3$ and the axis-angle rotation  ${\boldsymbol {\omega }}_{PN}\in \mathbb{R}^3$.
\end{enumerate}

The network architectures are based on the Monodepth2 implementation~\cite{Monodepth2} and use PyTorch~\cite{paszke2019pytorch}. Both networks encoders are initialized with pre-trained weights on the ImageNet dataset~\cite{deng2009imagenet}.

\subsection{Experimental settings}
The images are resized to $640\times192$ before entering the network. During training, we sample with repetition 2000 images for each epoch. We use standard color image augmentation by slightly changing saturation, brightness, contrast, and hue as in~\cite{Monodepth2}, and horizontal flipping. For the optimization, we use Adam~\cite{kingma2014adam} with parameters $\beta_1=0.9$ and $\beta_2=0.999$, and a learning rate $lr=10^{-4}$ . We half the learning rate when the loss does not decrease for 10 epochs. We keep the training until convergence of the loss or for at most 1800 epochs. The \textit{depth smoothness loss}, \textit{depth consistency loss}, and \textit{residual rotation loss} weighing factors are 10$^{-3}$, $5\times 10^{-1}$, and 1, respectively.

\subsection{KITTI Results}

\input{table_figures/kitti_res_tab}

\input{table_figures/kitti_traj_fig}

\input{table_figures/kitti_traj_test}

We evaluate our visual odometry network on KITTI odometry dataset~\cite{KITTI}. To this aim, we use the sequences from 0 to 8 for training and the sequences 9 and 10 for testing. Furthermore, we use the tool provided by the author of DF-VO~\cite{DF-VO} to make sure we apply the same criteria for evaluation. Notably, we evaluate with the ``7DoF alignment'' setting that computes the similarity transform that best aligns the predicted trajectory with the ground truth using the Umeyama algorithm~\cite{umeyama1991least}.

In Figure~\ref{fig:kitti-traj}, we show the plots of the trajectories for the training sequences predicted by our two models and the ground-truth poses. By comparing these with the testing sequences displayed in Figure~\ref{fig:kitti-traj-test}, we can appreciate the generalization capability of the neural network to unseen sequences even if KITTI contains images from similar scenarios. Then, in Table~\ref{table:kitti_benchmark}, we compare our results with two pure geometrical approaches, ORB-SLAM~\cite{ORB_SLAM} and VISO2~\cite{VISO2}, two Unsupervised Networks methods, SfM-Learner~\cite{Zhou17SFMlearner} and SC-SfMLearner~\cite{Bian19}, and with the hybrid approach DF-VO~\cite{DF-VO}. For the evaluation we use data from~\cite{DF-VO}. We note that the reported results for~\cite{Bian19} are slightly different from the one in the paper and may refer to training with additional data. For our evaluation, we select those works that use only monocular image sequences during training and evaluation phases as RAUM-VO because stereo image pairs give an unfair advantage to the depth reconstruction and, consequently, to the pose estimation, as documented in the literature~\cite{Monodepth2}. Another condition for the evaluation regards the architectures of the depth and pose networks. Therefore, we selected methods in the learned categories that use comparable, if not equal, deep networks. Unfortunately, this is one element of discrepancy among the works in the literature of unsupervised pose and depth estimation, and that has to be taken into account when making comparisons.

While RAUM-VO does not surpass DF-VO performances in many sequences, his accuracy is comparable while being more efficient. Because DF-VO is one of the most promising hybrid approaches using monocular images for the VO, in Section~\ref{sec:df-vo}, we examine the differences and advantages of our method in more detail. Regarding traditional methods, the average error of RAUM-VO is generally lower except only for the $r_{err}$ metric computed on ORB-SLAM. However, unlike ORB-SLAM, we do not apply local BA. Regarding the Unsupervised Pose Networks category, the proposed RAUM-VO proves to reduce the error effectively with the proposed rotation adjustment step.
Through the link~\footnote{\url{https://youtu.be/4woTiJRCrUI}}, we provide a video that shows the depth map predictions for all the KITTI sequences.

%% file: table_figures/kitti_res_tab.tex
\begin{table*} [t] 
\caption{ \textbf{Odometry quantitative evaluation}. Result obtained on KITTI Odometry Seq. 00-10. Data is retrieved from ~\cite{DF-VO}. Best results are highlighted in \textbf{bold}, second bests with an \underline{underline}.
}
\begin{center}
\resizebox{\columnwidth}{!}{%
\begin{tabular}{ c | c | c | c c c c c c c c c c c | c | c }
\toprule
Category & Method & Metric & 00 & 01 & 02 & 03 & 04 & 05 & 06 & 07 & 08 & 09 & 10 & Train Avg. Err. & Tot. Avg. Err.\\
\midrule

\multirow{10}*{Geometric} &
\multirow{5}*{\shortstack[1]{ORB-SLAM2\cite{ORB_SLAM}\\ (w/o LC)}} & $t_{err}$ & 11.43 & 107.57 & 10.34 & \textbf{0.97} & \textbf{1.30} & 9.04 & 14.56 & 9.77 & 11.46 & 9.30 & 2.57 & 19.604 & 17.119\\ 
& & $r_{err}$ & \textbf{0.58} & 0.89 & \textbf{0.26} & \textbf{0.19} & \textbf{0.27} & \underline{0.26} & \textbf{0.26} & \underline{0.36} & \textbf{0.28} & \underline{0.26} & \textbf{0.32} & \textbf{0.372}  & \textbf{0.357}\\ 
& & ATE & 40.65 & 502.20 & 47.82 & \textbf{0.94} & \textbf{1.30} & 29.95 & 40.82 & 16.04 & 43.09 & 38.77 & \underline{5.42} & 80.312  & 69.727\\ 
& & RPE (m) & 0.169 & 2.970 & 0.172 & 0.031 & 0.078 & 0.140 & 0.237 & 0.105 & 0.192 & 0.128 & \underline{0.045} & 0.455 &0.388\\ 
& & RPE ($^\circ$) &  0.079 & 0.098 & 0.072 & 0.055 & 0.079 & 0.058 & 0.055 & 0.047 & 0.061 & 0.061 & 0.065 & 0.067&0.066 \\ 

\cline{2-16} 

& \multirow{5}*{\shortstack[1]{VISO2\cite{VISO2}}} & $t_{err}$   & 10.53 & 61.36 & 18.71 & 30.21 & 34.05 & 13.16 & 17.69 & 10.80 & 13.85 & 18.06 & 26.10 & 23.373 & 23.138\\ 
& & $r_{err}$ & 2.73 & 7.68 & 1.19 & 2.21 & 1.78 & 3.65 & 1.93 & 4.67 & 2.52 & 1.25 & 3.26 & 3.151 &2.988\\ 
& & ATE & 79.24 & 494.60 & 70.13 & 52.36 & 38.33 & 66.75 & 40.72 & 18.32 & 61.49 & 52.62 & 57.25 &  102.438 & 93.801\\ 
& & RPE (m) & 0.221 & 1.413 & 0.318 & 0.226 & 0.496 & 0.213 & 0.343 & 0.191 & 0.234 & 0.284 & 0.442 & 0.406  &0.398 \\ 
& & RPE ($^\circ$) &  0.141 & 0.432 & 0.108 & 0.157 & 0.103 & 0.131 & 0.118 & 0.176 & 0.128 & 0.125 & 0.154 &0.166  &0.161 \\ 
\midrule

\multirow{15}*{Unsupervised} & \multirow{5}*{\shortstack[1]{SfM-Learner\cite{Zhou17SFMlearner}}} & $t_{err}$   &  21.32 & 22.41 & 24.10 & 12.56 & 4.32 & 12.99 & 15.55 & 12.61 & 10.66 & 11.32 & 15.25 & 15.169 & 14.826 \\ 
& & $r_{err}$ & 6.19 & 2.79 & 4.18 & 4.52 & 3.28 & 4.66 & 5.58 & 6.31 & 3.75 & 4.07 & 4.06 &4.584 &4.490 \\ 
& & ATE & 104.87 & 109.61& 185.43 & 8.42 & 3.10 & 60.89 & 52.19 & 20.12 & 30.97 & 26.93 & 24.09 &  63.956  & 56.965\\ 
& & RPE (m) & 0.282 & 0.660 & 0.365 & 0.077 & 0.125 & 0.158 & 0.151 & 0.081 & 0.122 & 0.103 & 0.118 & 0.225 &0.204 \\ 
& & RPE ($^\circ$) &  0.227 & 0.133 & 0.172 & 0.158 & 0.108 & 0.153 & 0.119 & 0.181 & 0.152 & 0.159 & 0.171 & 0.156  &0.158\\ 

\cline{2-16} 

& \multirow{5}*{\shortstack[1]{SC-SfMLearner\cite{Bian19}}} & $t_{err}$   & 11.01 & 27.09 & 6.74 & 9.22 & 4.22 & 6.70 & 5.36 & 8.29 & 8.11 & 7.64 & 10.74 & 9.638  & 9.556 \\
& & $r_{err}$ & 3.39 & 1.31 & 1.96 & 4.93 & 2.01 & 2.38 & 1.65 & 4.53 & 2.61 & 2.19 & 4.58 & 2.752  &2.867\\ 
& & ATE & 93.04 & 85.90 & 70.37 & 10.21 & 2.97 & 40.56 & 12.56 & 21.01 & 56.15 & 15.02 & 20.19 &43.641   & 38.907\\ 
& & RPE (m) & 0.139 & 0.888 & 0.092 & 0.059 & 0.073 & 0.070 & 0.069 & 0.075 & 0.085 & 0.095 & 0.105 & 0.172&0.159 \\ 
& & RPE ($^\circ$) & 0.129 & 0.075 & 0.087 & 0.068 & 0.055 & 0.069 & 0.066 & 0.074 & 0.074 & 0.102 & 0.107 &0.077 &0.082 \\ 

\cline{2-16} 

 & \multirow{5}*{\textbf{\shortstack[1]{Simple-Mono-VO\\(Ours)}}} & $t_{err}$ & 9.365 &   \underline{8.920} &   6.830 &  3.697 &  2.570 &   4.964 &  3.138 &  3.568 &   7.125 &  13.625 &  11.131 & \underline{5.575}  &6.812\\
& & $r_{err}$ &   2.840 &   \underline{0.562} &   1.582 &  2.478 &  0.566 &   2.083 &  0.959 &  1.866 &   2.608 &   3.146 &   4.784 & 1.727  &2.134\\
& & ATE &  94.949 &  \underline{30.004} &  83.155 &  4.112 &  2.377 &  30.227 &  8.726 &  8.872 &  59.887 &  66.591 &  18.792 & 35.812  & 37.063 \\
& & RPE (m) &   0.090 &   \underline{0.304} &   0.087 &  0.037 &  0.055 &   0.041 &  0.051 &  0.044 &   0.074 &   0.166 &   0.077 &  \underline{0.087} & 0.093 \\
& & RPE ($^\circ$) &   0.072 &   \textbf{0.042} &   0.057 &  \underline{0.048} &  0.036 &   0.049 &  \underline{0.040} &  \underline{0.048} &   0.052 &   0.067 &   0.083 & \underline{0.049} &0.054 \\
\midrule

\multirow{10}*{Hybrid} & \multirow{5}*{\shortstack[1]{DF-VO\cite{DF-VO}\\(Mono)}} & $t_{err}$  & \textbf{2.33} & 39.46 & \underline{3.24} & \underline{2.21} & \underline{1.43} & \textbf{1.09} & \textbf{1.15} & \textbf{0.63} & \textbf{2.18} & \textbf{2.40} & \textbf{1.82} & 5.969 & \underline{5.267} \\ 
& & $r_{err}$ & \underline{0.63} & \textbf{0.50} & \underline{0.49} & \underline{0.38} & \underline{0.30} & \textbf{0.25} & \underline{0.39} & \textbf{0.29}& \underline{0.32} & \textbf{0.24} & \underline{0.38} &\underline{0.394} & \underline{0.379} \\ 
& & ATE & \textbf{14.45} & 117.40 & \underline{19.69} & \underline{1.00} & \underline{1.39} & \textbf{3.61} & \textbf{3.20} & \textbf{0.98} & \textbf{7.63} & \textbf{8.36} & \textbf{3.13} & \underline{18.817} & \underline{16.440}\\ 
& & RPE (m) & \textbf{0.039} & 1.554 & \underline{0.057} & \textbf{0.029} & \textbf{0.046} & \textbf{0.024} & \textbf{0.030} & \textbf{0.021} & \textbf{0.041} & \textbf{0.051} & \textbf{0.043} & 0.205 & \underline{0.176} \\ 
& & RPE ($^\circ$) &  \textbf{0.056} & \underline{0.049} & \textbf{0.045} & \textbf{0.038} & \textbf{0.029} & \textbf{0.035} & \textbf{0.029} & \textbf{0.030} & \textbf{0.037} & \textbf{0.036} & \textbf{0.043} & \textbf{0.039} & \textbf{0.039} \\ 

\cline{2-16} 

    & \multirow{5}*{\textbf{\shortstack[1]{RAUM-VO\\(Ours)}}} & $t_{err}$& \underline{2.548} &   \textbf{8.354} &   \textbf{2.578} &  3.217 &  2.860 &   \underline{3.045} &  \underline{3.033} &  \underline{2.390} &   \underline{3.632} &  \underline{2.927} &   \underline{5.843} & \textbf{3.517}  & \textbf{3.675}\\
& & $r_{err}$ &   0.775 &   0.868 &   0.582 &  1.334 &  0.645 &   1.153 &  0.837 &  1.037 &   1.074 &  0.318 &   0.683 & 0.923 &0.846 \\
& & ATE &  \underline{16.272} &  \textbf{23.748} &  \textbf{16.139} &  2.602 &  2.283 &  \underline{17.470} &  \underline{9.234} &  \underline{2.164} &  \underline{16.303} &  \underline{8.664} &  12.297 & \textbf{11.802} & \textbf{11.561} \\
& & RPE (m) &  \underline{0.040} &   \textbf{0.257} &   \textbf{0.050} &  \underline{0.030} &  \underline{0.052} &   \underline{0.038} &  \underline{0.046} &  \underline{0.028} &   \underline{0.053} &  \underline{0.068} &   0.078 & \textbf{0.066} &  \textbf{0.067} \\
& & RPE ($^\circ$) &  \underline{0.059} &   0.062 &   \underline{0.048} &  \underline{0.048} &  \underline{0.035} &   \underline{0.044} &  0.042 &  0.058 &   \underline{0.045} &  \underline{0.042} &   \underline{0.051} & \underline{0.049} & \underline{0.049} \\

\bottomrule

\end{tabular}
}
\end{center}
\vspace{-10pt}
\label{table:kitti_benchmark}
\end{table*}

%% file: table_figures/kitti_traj_fig.tex
\begin{figure*}[!ht]
\captionsetup{justification=centering, belowskip=1pt}

  \centering
  \begin{subfigure}{.32\columnwidth}
    \centering\includegraphics[width=\linewidth]{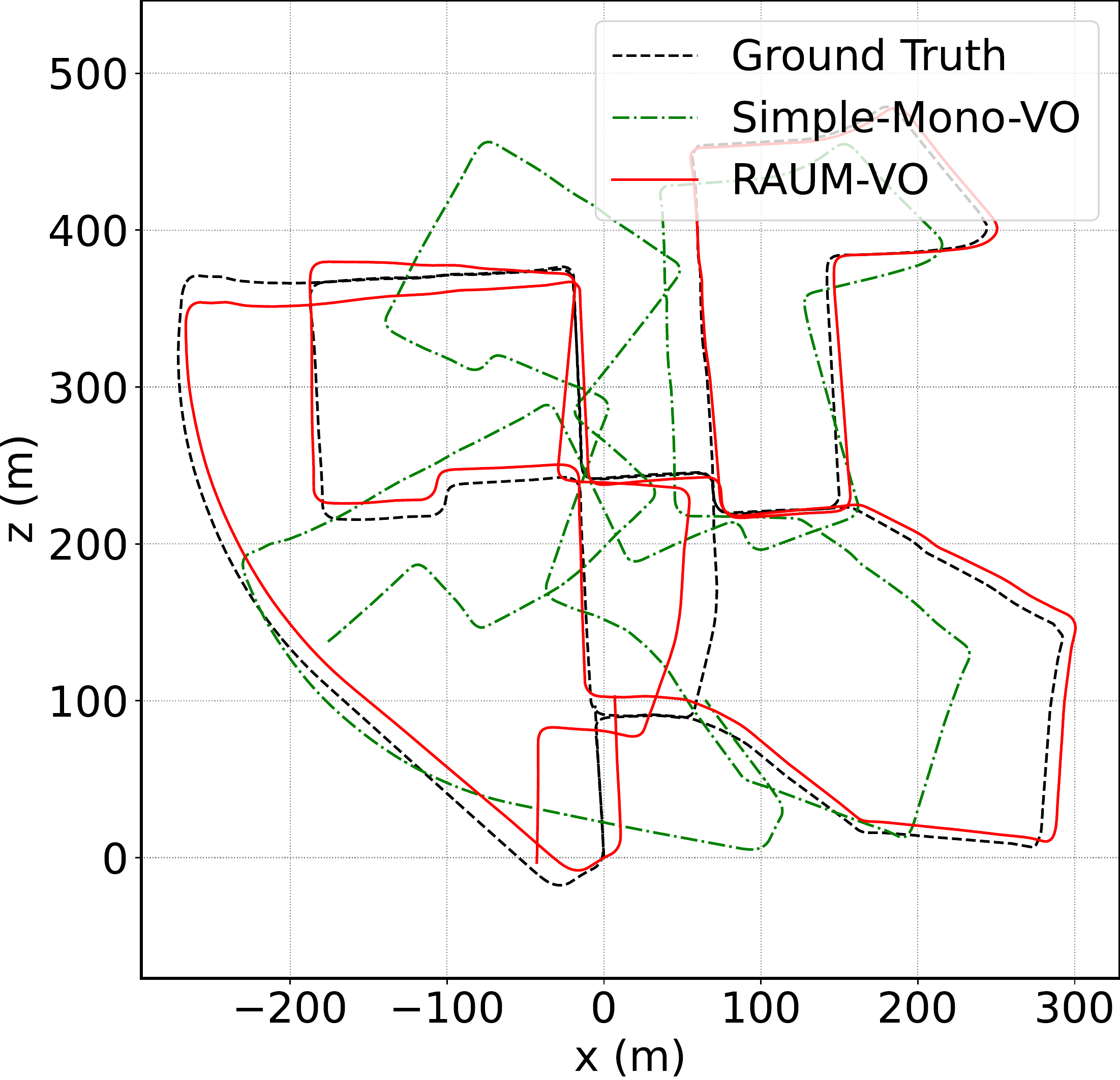}
    \caption{Seq. 00}
  \end{subfigure}
    \begin{subfigure}{.32\columnwidth}
    \centering\includegraphics[width=\linewidth]{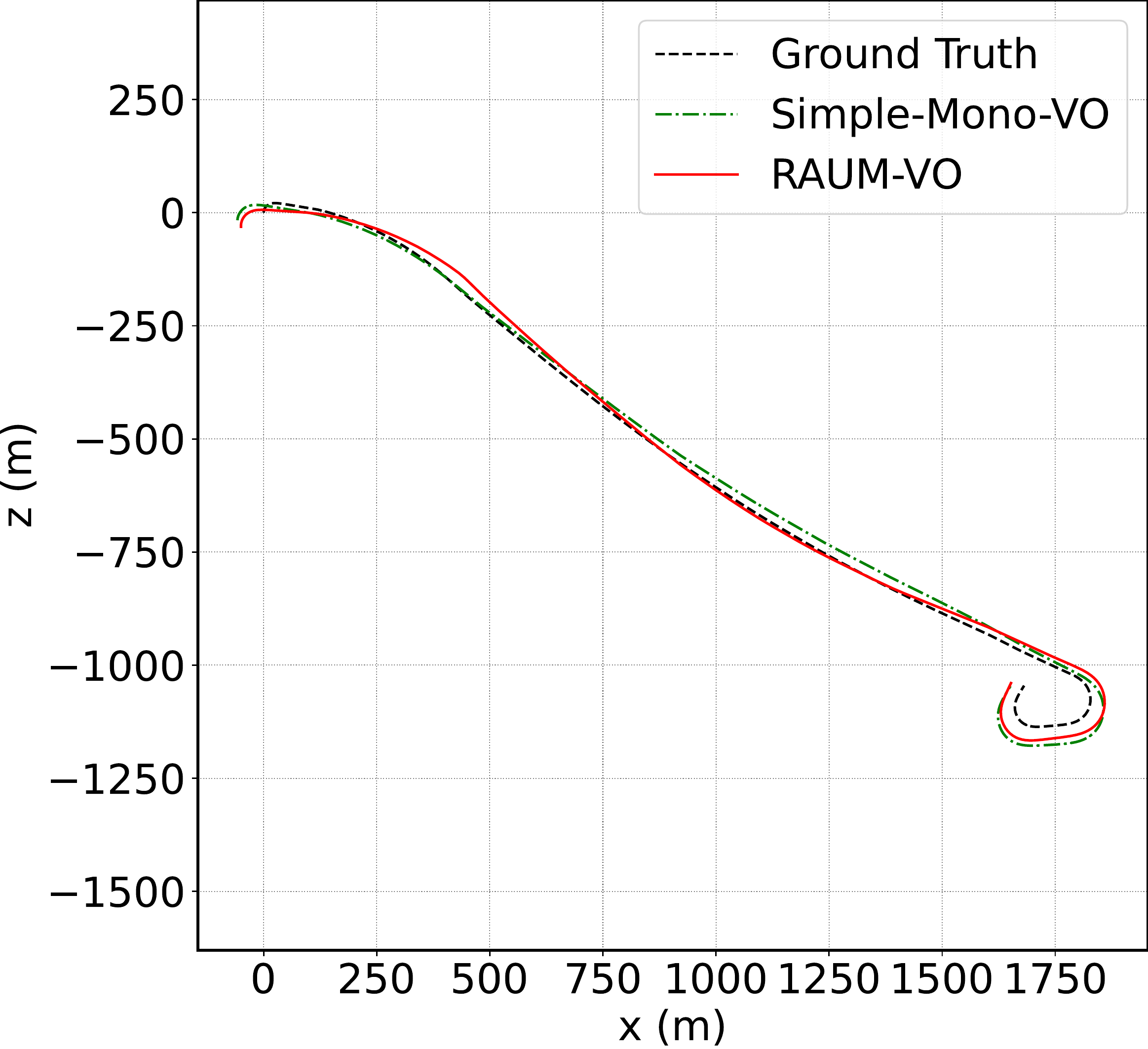}
    \caption{Seq. 01}
  \end{subfigure}
  \begin{subfigure}{.32\columnwidth}
    \centering\includegraphics[width=\linewidth]{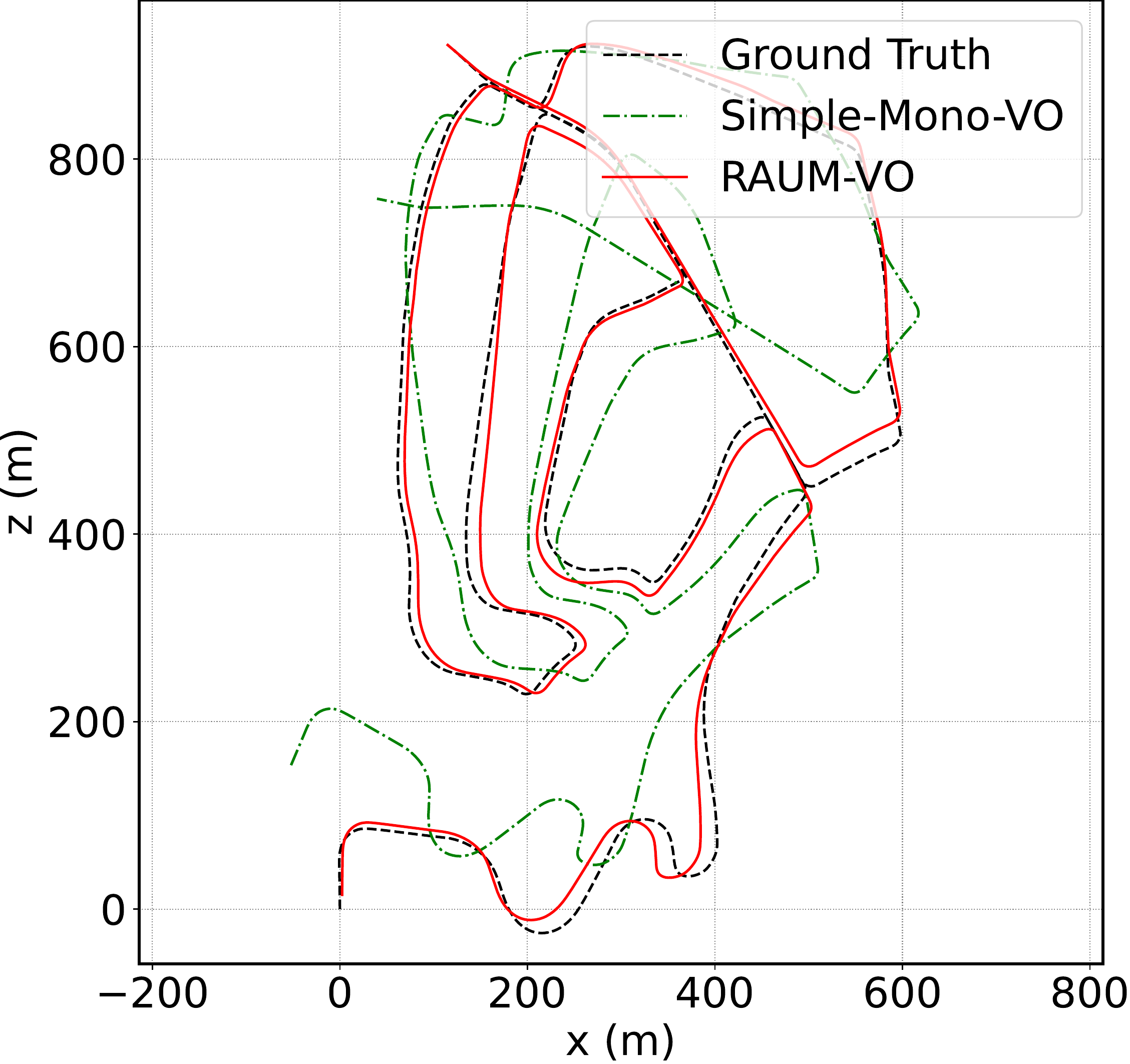}
    \caption{Seq. 02}
  \end{subfigure}
 
  \begin{subfigure}{.32\columnwidth}
    \centering\includegraphics[width=\linewidth]{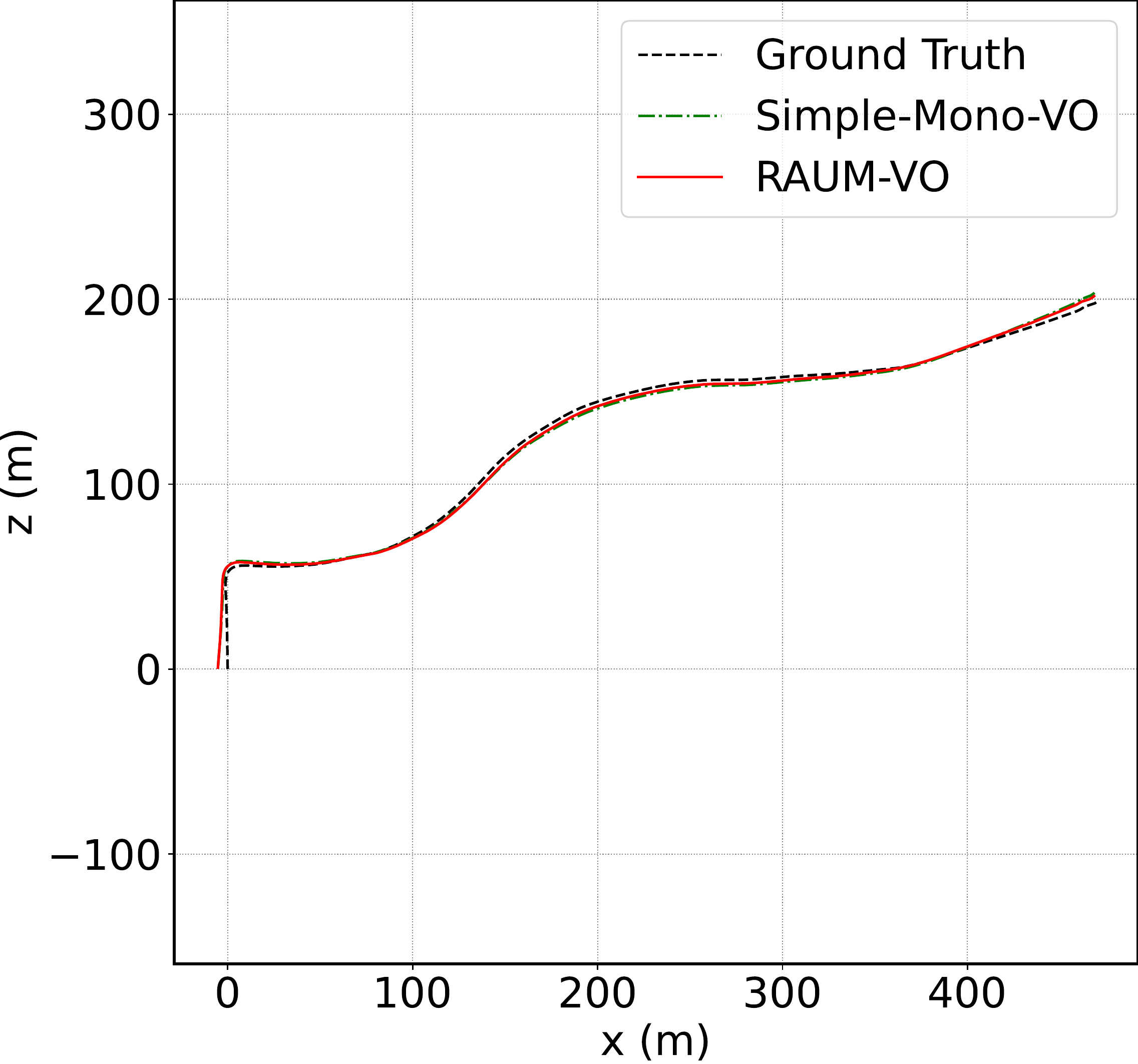}
    \caption{Seq. 03}
  \end{subfigure}
  \begin{subfigure}{.32\columnwidth}
    \centering\includegraphics[width=\linewidth]{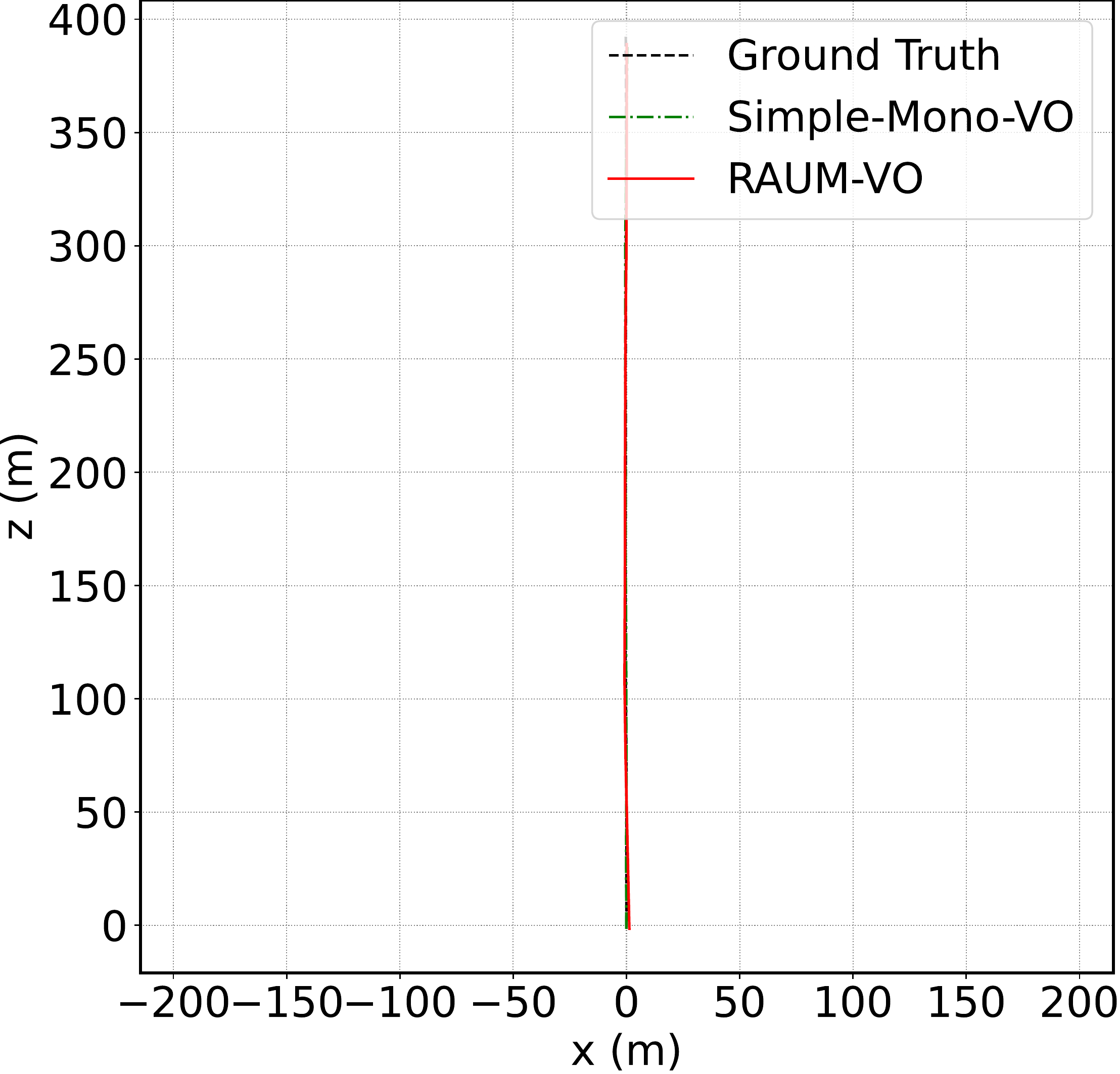}
    \caption{Seq. 04}
  \end{subfigure}
  \begin{subfigure}{.32\columnwidth}
    \centering\includegraphics[width=\linewidth]{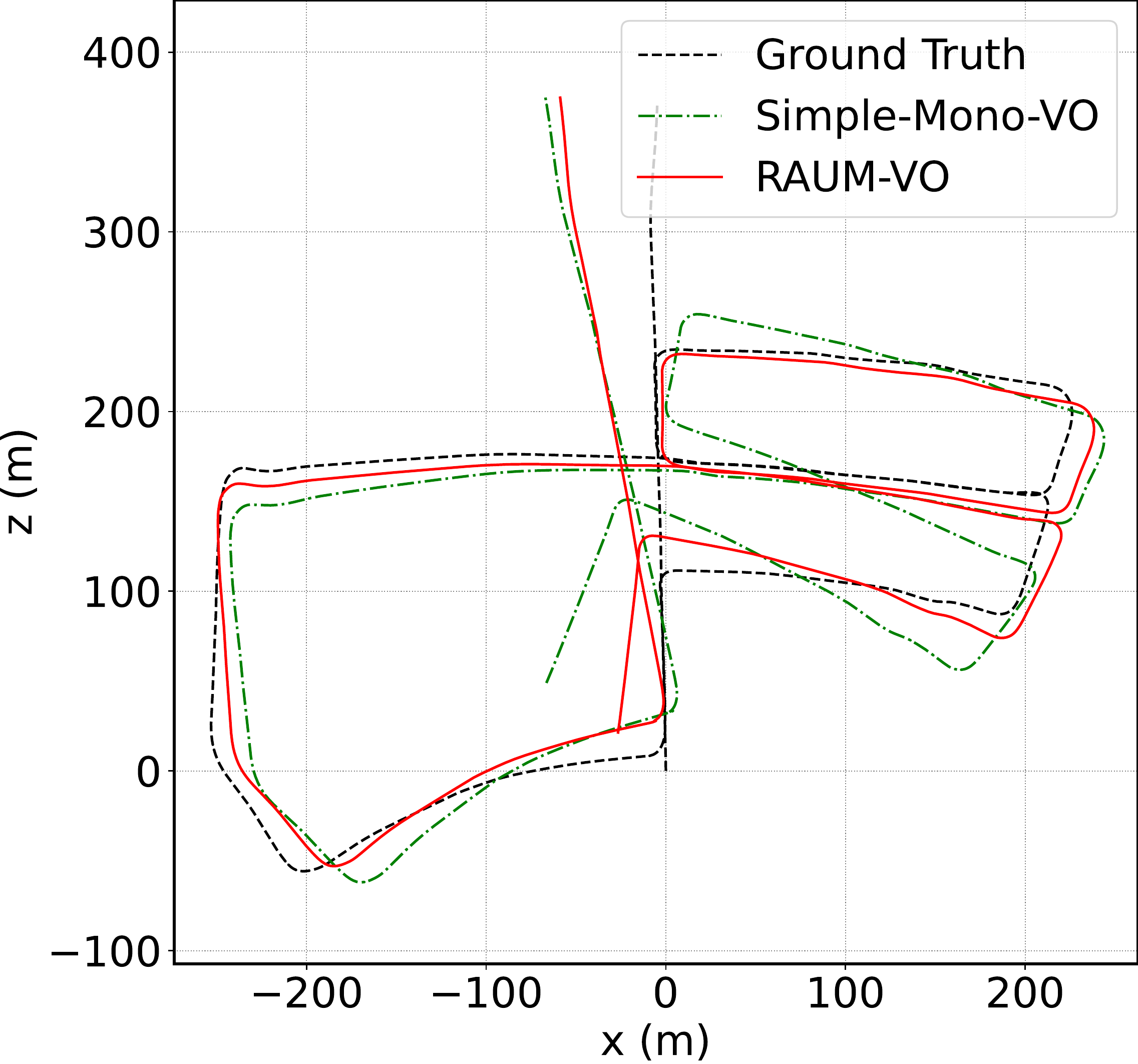}
    \caption{Seq. 05}
  \end{subfigure}
 
  \begin{subfigure}{.32\columnwidth}
    \centering\includegraphics[width=\linewidth]{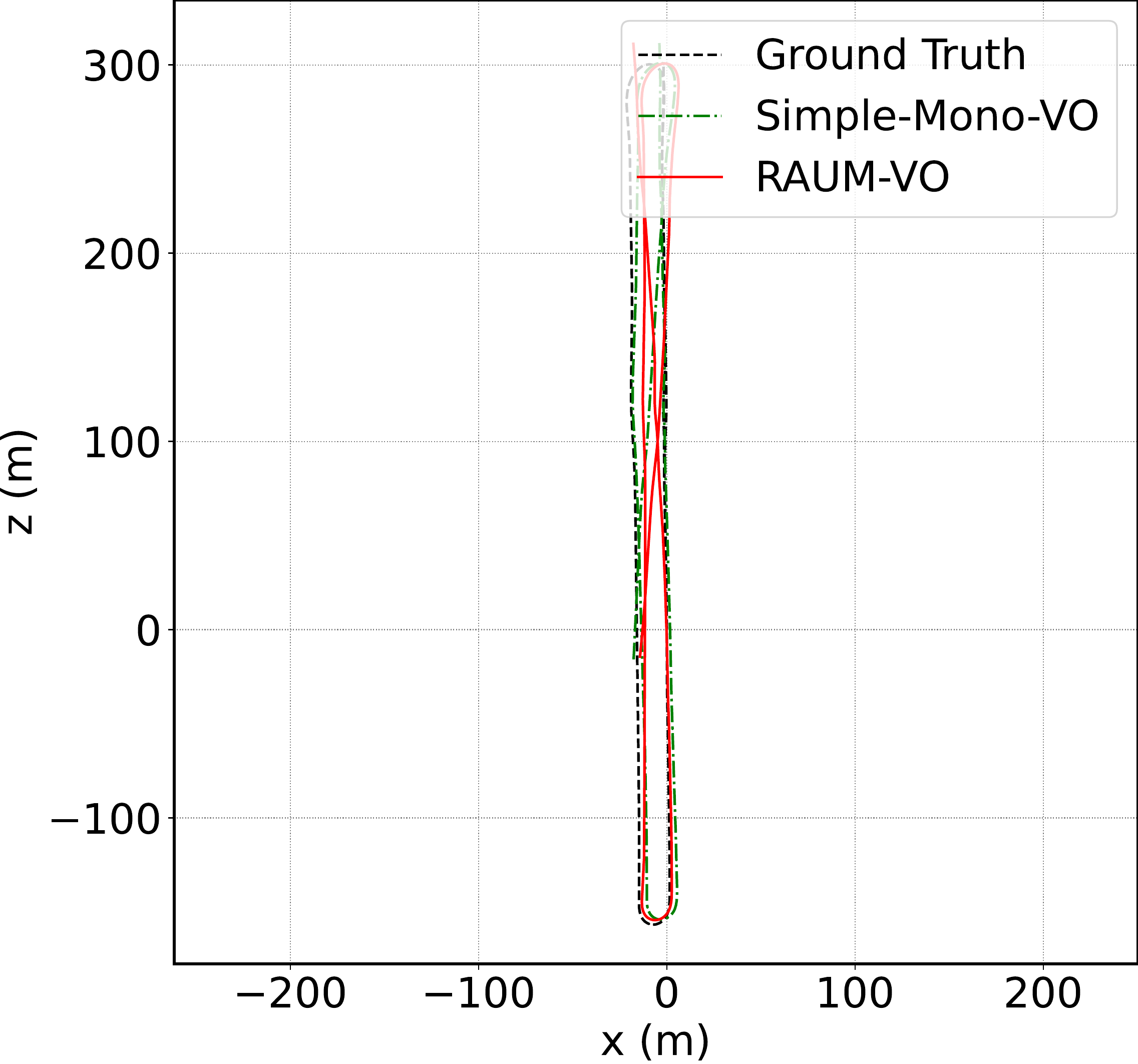}
    \caption{Seq. 06}
  \end{subfigure}
  \begin{subfigure}{.32\columnwidth}
    \centering\includegraphics[width=\linewidth]{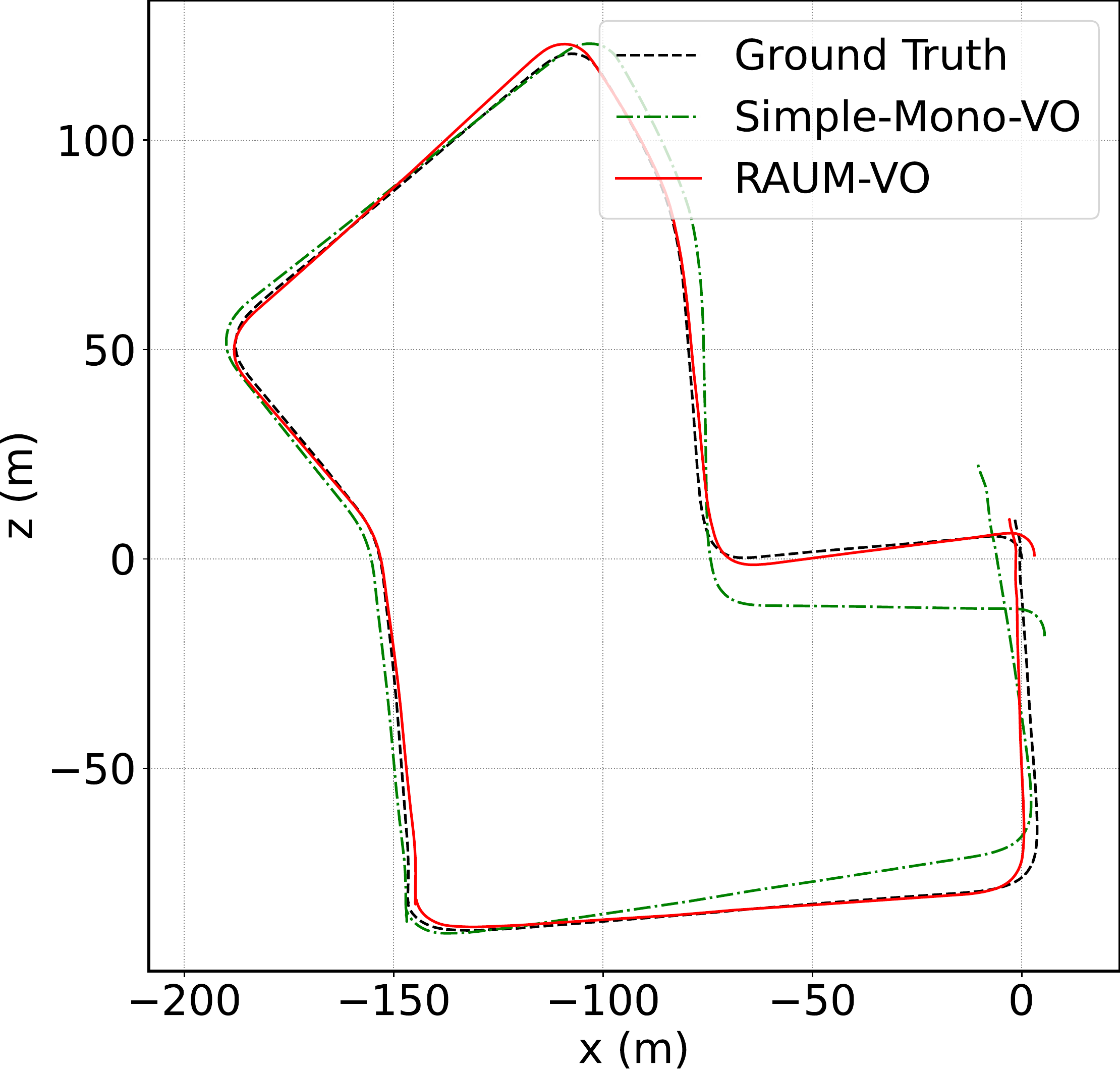}
    \caption{Seq. 07}
  \end{subfigure}
  \begin{subfigure}{.32\columnwidth}
    \centering\includegraphics[width=\linewidth]{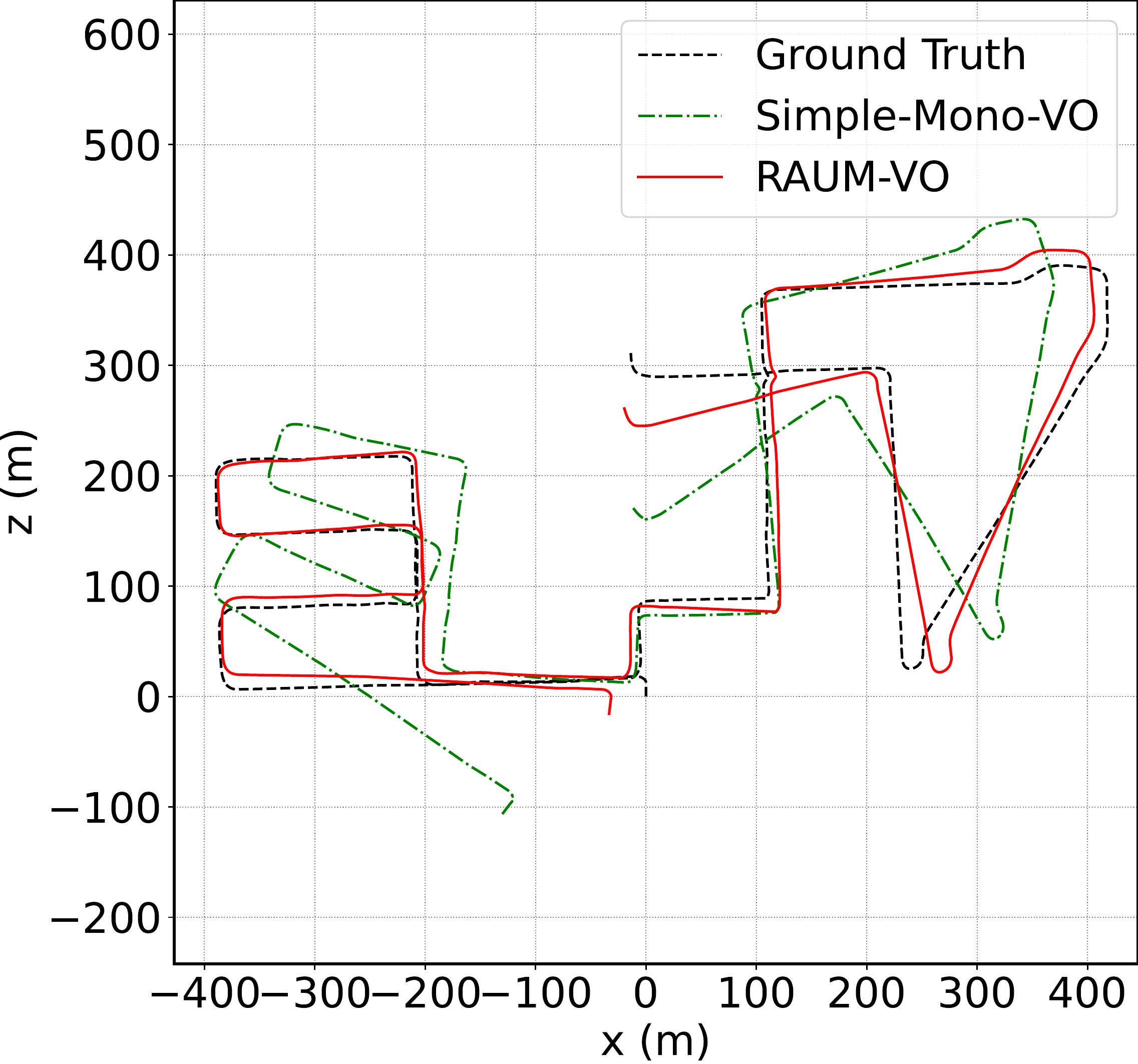}
    \caption{Seq. 08}
  \end{subfigure}

  \caption{\textbf{KITTI Train trajectories}. Estimated trajectories for the KITTI Odometry sequences from 00 to 08.Poses are given in camera frame. Thus, positive $x$ means right direction and positive $z$ means forward. Best viewed in colors.}
   \label{fig:kitti-traj}
\end{figure*}

%% file: table_figures/kitti_traj_test.tex
\begin{figure*}[!ht]
\captionsetup{justification=centering}
\centering
    \begin{subfigure}{.40\columnwidth}
    \centering\includegraphics[width=\linewidth]{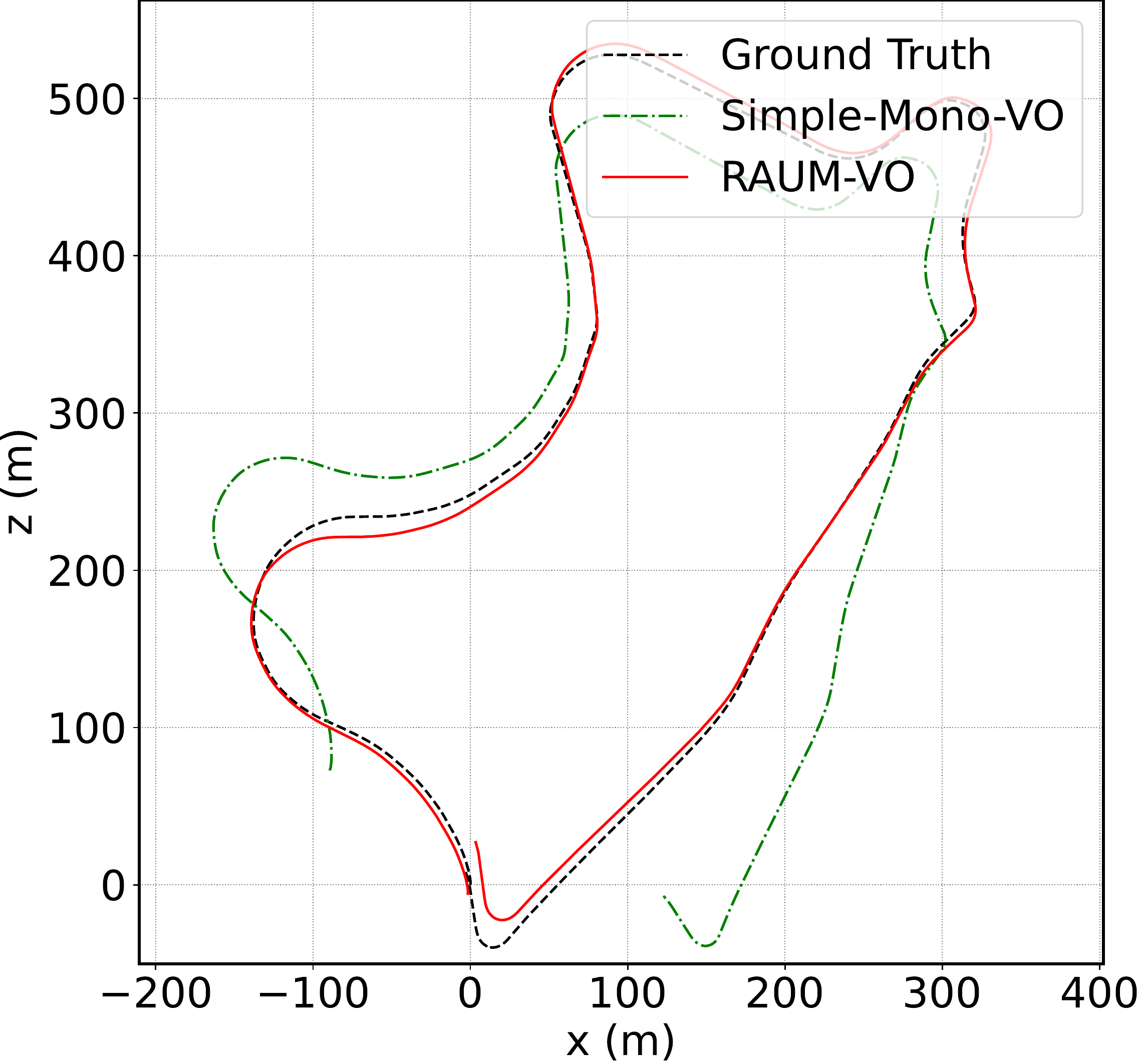}
    \caption{Seq. 09}
  \end{subfigure}
  \hspace{0.08\columnwidth}
    \begin{subfigure}{.40\columnwidth}
    \centering\includegraphics[width=\linewidth]{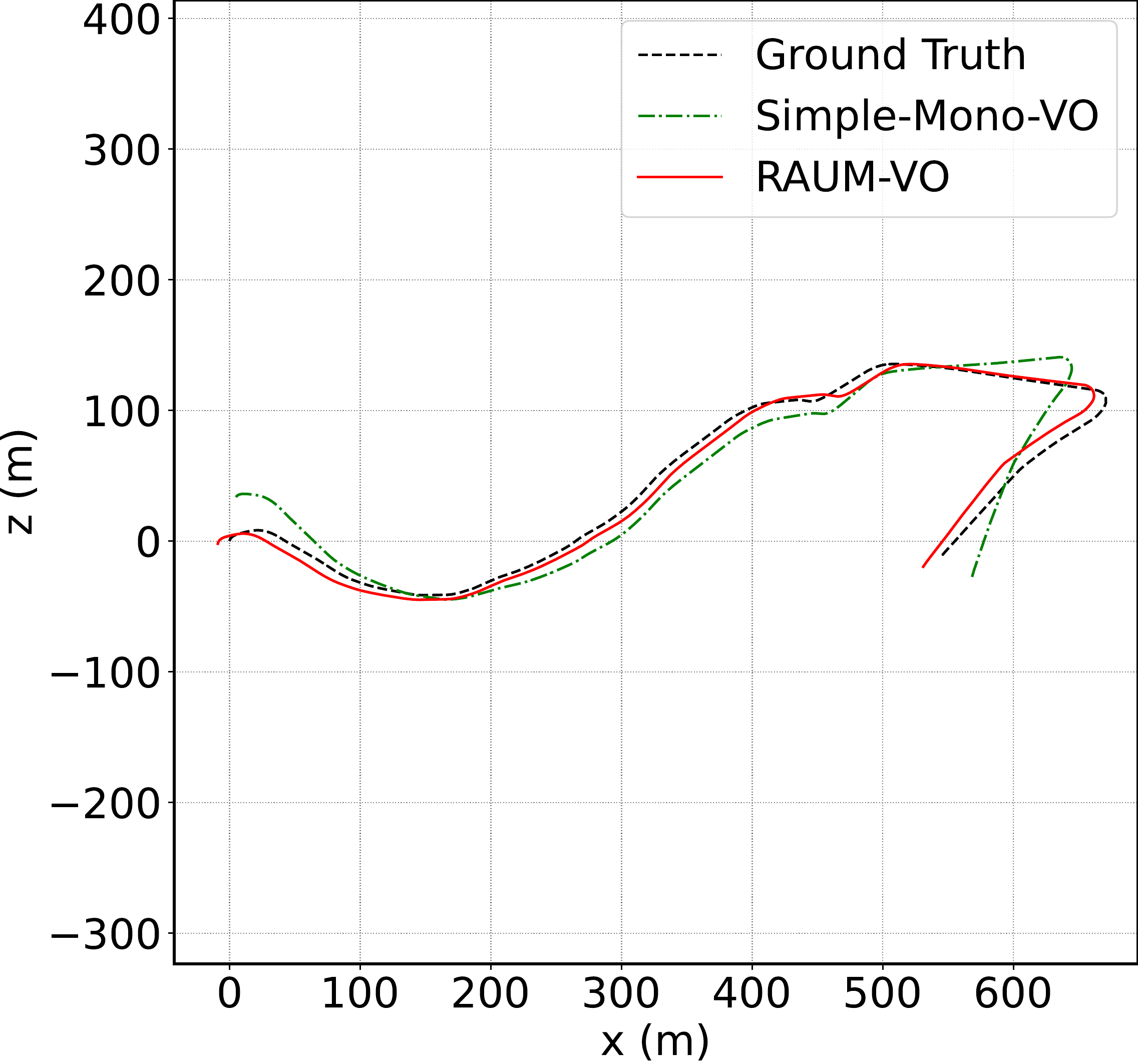}
    \caption{Seq. 10}
  \end{subfigure}
  \caption{\textbf{KITTI Test trajectories}. Estimated trajectories for the KITTI Odometry sequences 09 and 10. Poses are given in camera frame. Thus, positive $x$ means right direction and positive $z$ means forward. Best viewed in colors.}
   \label{fig:kitti-traj-test}
\end{figure*}

%% file: latex/content/discussion.tex
\section{Discussion}

Herein, we discuss and analyze the characteristics of RAUM-VO. First, in Section~\ref{sec:gc}, we consider the rotational and translational components of the pose error separately to argue that the rotations offer a larger space to decrease the Absolute Trajectory Error (ATE) shown in Table~\ref{table:kitti_benchmark}. In turn, this motivates the adoption of a specific measure to adjust the predicted rotations. Hence, we demonstrate how the pose network plays a valuable role in initializing the F2F solver. Lastly, in Section~\ref{sec:df-vo}, we speculate on the factor that contributes the most to the accuracy of DF-VO compared to our approach. 

\subsection{General Considerations}
\label{sec:gc}
\input{table_figures/gt_test_tab}
In Table~\ref{table:gttest}, we show that by modifying the Simple-Mono-VO predictions using the ground-truth of either the translation or the rotation, there is a larger margin for improvement enclosed in the current rotation estimates than in the translational component of the error. We presume that this behavior is because we optimize translations directly on their vector space contrary to the rotations. The manifold of rotations, \textit{special orthogonal} group $SO(3)$ only locally resembles a Euclidean topology~\cite{lee2013smooth} and needs intermediate representations to enable the optimization with gradient descent methods. As such, the axis-angles are a many-to-one mapping with $SO(3)$, and alternative representations may be easier to approximate with a neural network~\cite{zhou2019continuity}. Also, the linear distance metric between translation vectors is easier to approximate than the non-linear counterparts for the $SO(3)$ group~\cite{huynh2009metrics}. Nevertheless, the rotation provided by the Pose Network is a better initialization point for the F2F than the identity or constant motion assumption. The results of the different types of initialization are visible in Table~\ref{table:rotinit}. By this, the Pose Network predicted rotations are always the best option for initializing the F2F solver and are paired only by constant motion assumption in some cases.

\input{table_figures/rot_int_tab}

Then, we suggest that the Pose Network can regress the motion even in difficult motion situations, assuming that the Depth Network has learned a valid geometric structure. The pose and depth outcomes are strongly entangled due to their joint training, even if produced by separate networks. But, more precisely, we note that the performance of one component may be restricted by the other. While this may seem a trivial conclusion, it is necessary to clarify the limitations of this approach and bring us to the last reflection. We evaluate the odometry poses obtained by PnP combined with the Depth Network to prove our argument. To this aim, we back-project to 3D coordinates the matches in one view frame, the same utilized for our RAUM-VO, by interpolating the depth map values with the bilinear sampler of STNs.

Consequently, we can apply PnP with RANSAC to estimate the two view motions for all the sequences. Remarkably, the outcome of PnP, on average, matches closely that of the Pose Network (see Table~\ref{table:pnp}), especially for the training sequences when we fix the rotation with F2F. This result aligns with those of, for example, DeepMatchVO~\cite{shen2019beyond} or DF-VO~\cite{DF-VO}, which do not obtain significantly better odometry results by leveraging PnP directly during the training or at the test time. Interestingly though, the combination of a PnP with the estimated depths works best for the test sequences indicating that this approach may generalize more.

\input{table_figures/pnp_tab}

\subsection{Comparison with DF-VO}
 \label{sec:df-vo}
We can ascribe most probably the success of DF-VO to an accurately trained optical flow, which provides a significantly higher number of precise matches in the order of thousands. Still, these correspondences are specific to the scenario they train the optical flow network. Conversely, the 2D features detected by Superpoint are fast to compute, distinctly identified, repeatable, and, more importantly, sparser (a few hundred). Therefore, we note that the optical flow network can hardly reach the generalization capability of a dedicated feature extraction network. Additionally, due to dense but noisy correspondences, DF-VO needs to search iterative the best fit model, \eg, based on the number of inliers, and decide between Essential or Homography motion model with multiple RANSAC routines. While this approach accurately describes the two-view motion of the KITTI sequences, it turns out to be computationally expensive. Instead, RAUM-VO uses all the matches found by Superglue for solving the eigenvalue minimization problem of F2F only once, adding minimal overhead to the Pose Network run-time. Thus, we remove the need for repeated samples of the correspondences and avoid the numerous estimation of Homography and Essential matrices with the related model selection strategy. Therefore, we resort to the output of the Pose Network and a single model-free rotation adjustment step, which is comparably a more efficient approach.


 Furthermore, another potential determining factor of success is the depth scale consistency. DF-VO considers the depth maps as a source of multiple hypotheses for the translation vector scale. Thus, we can presume that the disparities jointly learned with the optical flow have a higher degree of long-term scale consistency and structure accuracy. In this way, the DF-VO scale alignment procedure can recover the best norm for the translation vector, which the employed Nister 5-point~\cite{Nister5points} algorithm delivers only up to a scale factor. Also, the depth consistency loss may not be as effective as the consistency loss between rigid motion and optical flow in maintaining a unique long-term scale factor.

Consequently, for evaluating our depth scale consistency, we applied a scale alignment procedure similar to DF-VO for scaling the translation solutions obtained from F2F and Essential matrix, using the implementations of OpenGV~\cite{kneip2014opengv} and OpenCV, respectively. Notably, we pick the Essential matrix with the most inliers after ten iterations, sampling each time 20$\%$ of the matches and estimating it using RANSAC with threshold $10^{-3}$. Next, we triangulate the 2D correspondences and keep only those that pass the cheirality check. Finally, we sample for ten times $80\%$ of the triangulated points $\mathbf{X}_t$ and fit a linear model with RANSAC:
 \begin{equation}
     \mathbf{Y}_d = s\,\mathbf{X}_t
 \end{equation}
 to find the coefficient $s$ that maps $\mathbf{X}_t$ to $\mathbf{Y}_d$, which is the set of 3D points obtained by projecting the matches with the estimated depths. Finally, we take the scale $s$ that has the minimum $\delta=\|1-s\|$. We fallback to the Pose Network estimated translation only if less than $51\%$ of matches do not pass the cheirality check or if $\delta > 5\times10^{-1}$. We accept the F2F or Essential matrix translation in the $93-97\%$ of the cases with these loose constraints. We present the result of this test in Table~\ref{table:scale_align}. Still, we could not obtain a better translation than the Pose Network's output. Besides, the multiple RANSAC routines and sampling matches from dense correspondences may grant a decisive advantage to DF-VO. We leave a deeper analysis to understand the factors at stake for future works.

\input{table_figures/scale_align_tab}

%% file: table_figures/gt_test_tab.tex
\begin{table} [htpb] 
\caption{The table shows an insight on the possible margins for improvement in the pose predictions coming from unsupervised methods. Hence, we substitute alternately the ground-truth translations and rotations to the Pose Network estimates. We show the variation on the relevant metrics for the KITTI test sequences 9 and 10.}
\begin{center}
\resizebox{.4\columnwidth}{!}{%
\begin{tabular}{ c | c | c c }
\toprule
    & Metrics  &   09 &      10  \\
   
\midrule
 \multirow{5}*{\textbf{\shortstack[1]{Ours \\Simple-Mono-VO}}} & $t_{err}$ &  13.625 &  11.131\\
 & $r_{err}$ &  3.146 &   4.784 \\
& ATE &   66.591 &  18.792 \\
& RPE (m) &  0.166 &   0.077 \\
 & RPE ($^\circ$) & 0.067 &   0.083 \\
 
\midrule

\multirow{5}*{\shortstack[1]{Ground-Truth \\Translation}}  & $t_{err}$ &  13.325 & 11.409 \\
            & $r_{err}$ &   3.146 &  4.784 \\
            & ATE &  65.081 &  20.715  \\
            & RPE (m) &   0.162 &  0.028  \\
            & RPE ($^\circ$) &  0.067  & 0.083\\
            
\midrule
\multirow{5}*{\shortstack[1]{Ground-Truth \\Rotation}}  & $t_{err}$ &   3.029  & 6.038\\
            & $r_{err}$ &  0.010 &  0.014 \\
            & ATE &  9.026 & 12.894 \\
            & RPE (m) &   0.070 &  0.080   \\
            & RPE ($^\circ$) &   0.005 &  0.005\\

\bottomrule
\end{tabular}
}
\end{center}
\vspace{-10pt}
\label{table:gttest}
\end{table}

%% file: table_figures/rot_int_tab.tex
\begin{table*} [t] 
\caption{\textbf{F2F solver initialization}. Comparison of different initialization approaches for the Levenberg-Marquardt scheme that solves the frame-to-frame motion. Overall, the rotation from the Pose Network is the best, followed by a constant motion model.}
\begin{center}
\resizebox{\columnwidth}{!}{%
\begin{tabular}{ c | c | c c c c c c c c c c c| c | c }
\toprule
   Initialization & Metrics  &    00 &      01 &      02 &     03 &     04 &      05 &     06 &     07 &      08 &      09 &      10 &  Avg.Train &  Avg.All \\
\midrule
\multirow{5}*{Identity}  & $t_{err}$ &   6.192 &   8.023 &   5.888 &  3.919 &  2.860 &   7.659 &   9.100 &  10.969 &   5.402 &   3.851 &   9.475 &      6.668 &    6.667 \\
            & $r_{err}$ &   2.222 &   1.025 &   1.670 &  1.909 &  0.645 &   3.340 &   2.926 &   6.565 &   1.926 &   0.742 &   2.605 &      2.470 &    2.325 \\
            & ATE &  39.195 &  21.231 &  91.621 &  2.651 &  2.283 &  40.192 &  19.682 &  20.592 &  30.142 &  12.939 &  13.399 &     29.732 &   26.721 \\
            & RPE (m) &   0.040 &   0.259 &   0.060 &  0.030 &  0.052 &   0.039 &   0.046 &   0.036 &   0.052 &   0.069 &   0.077 &      0.068 &    0.069 \\
            & RPE ($^\circ$) &   0.100 &   0.101 &   0.072 &  0.082 &  0.035 &   0.083 &   0.059 &   0.158 &   0.067 &   0.070 &   0.088 &      0.084 &    0.083 \\

\midrule
\multirow{5}*{Constant Motion} & $t_{err}$&   6.062 &  12.009 &   5.823 &  6.606 &  2.860 &   5.877 &  3.033 &  2.481 &  19.533 &   3.255 &   5.843 &      7.143 &    6.671 \\
            & $r_{err}$&   2.128 &   1.833 &   1.728 &  3.119 &  0.645 &   2.105 &  0.837 &  1.150 &   7.772 &   0.862 &   0.683 &      2.368 &    2.078 \\
            & ATE &  58.308 &  49.099 &  79.710 &  6.678 &  2.283 &  29.920 &  9.234 &  2.258 &  99.024 &  11.190 &  12.297 &     37.390 &   32.727 \\
            & RPE (m) &   0.044 &   0.265 &   0.056 &  0.030 &  0.052 &   0.039 &  0.046 &  0.028 &   0.160 &   0.069 &   0.078 &      0.080 &    0.079 \\
            & RPE ($^\circ$) &   0.075 &   0.086 &   0.059 &  0.066 &  0.035 &   0.060 &  0.042 &  0.068 &   0.702 &   0.072 &   0.051 &      0.133 &    0.120 \\
\midrule

 \multirow{5}*{\shortstack[1]{Pose Network \\ (\textbf{RAUM-VO})}} & $t_{err}$& 2.548 &   8.354 &   2.578 &  3.217 &  2.860 &   3.045 &  3.033 &  2.390 &   3.632 &  2.927 &   5.843 & 3.517  & 3.675\\
 & $r_{err}$ &   0.775 &   0.868 &   0.582 &  1.334 &  0.645 &   1.153 &  0.837 &  1.037 &   1.074 &  0.318 &   0.683 & 0.923 & 0.846 \\
 & ATE &  16.272 &  23.748 &  16.139 &  2.602 &  2.283 &  17.470 &  9.234 &  2.164 &  16.303 &  8.664 &  12.297 & 11.802 & 11.561 \\
 & RPE (m) &   0.040 &   0.257 &   0.050 &  0.030 &  0.052 &   0.038 &  0.046 &  0.028 &   0.053 &  0.068 &   0.078 & 0.066 &  0.067 \\
 & RPE ($^\circ$) &   0.059 &   0.062 &   0.048 &  0.048 &  0.035 &   0.044 &  0.042 &  0.058 &   0.045 &  0.042 &   0.051 & 0.049 & 0.049\\
\bottomrule
\end{tabular}
}
\end{center}
\vspace{-10pt}
\label{table:rotinit}
\end{table*}

%% file: table_figures/pnp_tab.tex
\begin{table*} [t] 
\caption{\textbf{PnP vs Pose Network}. Comparison of the trajectory estimated by PnP combined with the Depth Network and the poses predicted by our trained network.}
\begin{center}
\resizebox{\columnwidth}{!}{%
\begin{tabular}{c | c | c c c c c c c c c c c| c | c  }
\toprule
   Poses Source & Metrics  &    00 &      01 &      02 &     03 &     04 &      05 &     06 &     07 &      08 &      09 &      10 &  Avg.Train &  Avg.All \\
 
\midrule
\multirow{5}*{\shortstack[1]{Pose Network\\ (Simple-Mono-VO)}} & $t_{err}$&   9.365 &   8.920 &   6.830 &  3.697 &  2.570 &   4.964 &  3.138 &  3.568 &   7.125 &  13.625 &  11.131 &      5.575 &    6.812 \\
             & $r_{err}$&   2.840 &   0.562 &   1.582 &  2.478 &  0.566 &   2.083 &  0.959 &  1.866 &   2.608 &   3.146 &   4.784 &      1.727 &    2.134 \\
             & ATE &  94.949 &  30.004 &  83.155 &  4.112 &  2.377 &  30.227 &  8.726 &  8.872 &  59.887 &  66.591 &  18.792 &     35.812 &   37.063 \\
             & RPE (m) &   0.090 &   0.304 &   0.087 &  0.037 &  0.055 &   0.041 &  0.051 &  0.044 &   0.074 &   0.166 &   0.077 &      0.087 &    0.093 \\
             & RPE ($^\circ$) &   0.072 &   0.042 &   0.057 &  0.048 &  0.036 &   0.049 &  0.040 &  0.048 &   0.052 &   0.067 &   0.083 &      0.049 &    0.054 \\
  
\midrule
\multirow{5}*{PnP} & $t_{err}$&   6.808 &  17.627 &   6.319 &  4.046 &  2.627 &   4.629 &  2.981 &  3.013 &   6.360 &   7.019 &  6.708 &      6.045 &    6.194 \\
             & $r_{err}$&   2.190 &   1.195 &   1.339 &  2.364 &  0.582 &   1.863 &  0.781 &  1.691 &   2.317 &   2.029 &  2.644 &      1.591 &    1.727 \\
             & ATE &  79.125 &  63.596 &  76.800 &  4.402 &  2.424 &  29.000 &  8.660 &  7.106 &  52.700 &  35.664 &  9.576 &     35.979 &   33.550 \\
             & RPE (m) &   0.061 &   0.636 &   0.086 &  0.033 &  0.055 &   0.039 &  0.049 &  0.040 &   0.067 &   0.082 &  0.073 &      0.118 &    0.111 \\
             & RPE ($^\circ$) &   0.060 &   0.057 &   0.049 &  0.042 &  0.029 &   0.039 &  0.032 &  0.036 &   0.043 &   0.068 &  0.085 &      0.043 &    0.049 \\

\midrule

\multirow{5}*{\shortstack[1]{F2F rotation w/ \\PnP translation}} & $t_{err}$&   2.796 &  15.552 &   2.775 &  3.482 &  3.123 &   3.008 &  3.164 &  2.373 &   3.876 &  3.072 &  4.343 &      4.461 &    4.324 \\
             & $r_{err}$&   0.775 &   0.868 &   0.582 &  1.334 &  0.645 &   1.146 &  0.837 &  0.861 &   1.074 &  0.318 &  0.683 &      0.902 &    0.829 \\
             & ATE &  17.662 &  41.782 &  15.194 &  2.342 &  2.459 &  17.203 &  9.451 &  3.983 &  16.741 &  8.288 &  8.909 &     14.091 &   13.092 \\
             & RPE (m) &   0.043 &   0.527 &   0.053 &  0.034 &  0.055 &   0.040 &  0.050 &  0.035 &   0.055 &  0.071 &  0.073 &      0.099 &    0.094 \\
             & RPE ($^\circ$) &   0.059 &   0.062 &   0.048 &  0.048 &  0.035 &   0.045 &  0.042 &  0.059 &   0.046 &  0.042 &  0.051 &      0.049 &    0.049 \\
\midrule

\multirow{5}*{\shortstack[1]{F2F rotation w/ \\Pose Network translation \\ (\textbf{RAUM-VO} w/o $L_r$)}} & $t_{err}$&   2.829 &   9.870 &   2.766 &  4.146 &  3.080 &   3.029 &  3.177 &  2.802 &   3.804 &  3.130 &   5.875 &      3.945 &    4.046 \\
             & $r_{err}$&   0.775 &   0.868 &   0.582 &  1.334 &  0.645 &   1.146 &  0.837 &  0.861 &   1.074 &  0.318 &   0.683 &      0.902 &    0.829 \\
             & ATE &  18.339 &  28.499 &  15.497 &  2.468 &  2.419 &  17.363 &  9.502 &  4.732 &  16.426 &  9.033 &  12.410 &     12.805 &   12.426 \\
             & RPE (m) &   0.043 &   0.307 &   0.053 &  0.037 &  0.055 &   0.041 &  0.051 &  0.036 &   0.056 &  0.070 &   0.079 &      0.075 &    0.075 \\
             & RPE ($^\circ$) &   0.059 &   0.062 &   0.048 &  0.048 &  0.035 &   0.045 &  0.042 &  0.059 &   0.046 &  0.042 &   0.051 &      0.049 &    0.049 \\

\bottomrule
\end{tabular}
}
\end{center}
\vspace{-10pt}
\label{table:pnp}
\end{table*}

%% file: table_figures/scale_align_tab.tex
\begin{table} [htpb] 
\caption{\textbf{Scale alignment}. Results of the scale alignment procedure applied to the translation vector from the F2F and the Essential matrix estimated motions.}
\begin{center}
\resizebox{.4\columnwidth}{!}{%
\begin{tabular}{ c | c | c c }
\toprule
    & Metrics  &   09 &      10  \\
   
\midrule
 \multirow{3}*{\shortstack[1]{F2F \\translation}} & $t_{err}$ & 4.14 & 5.68\\
& ATE & 12.91 & 11.67  \\
& RPE (m) & 0.114 & 0.091 \\

\midrule

\multirow{3}*{\shortstack[1]{Essential matrix \\translation}}  & $t_{err}$ & 4.02  & 5.99 \\
& ATE & 11.77  & 12.42  \\
& RPE (m) & 0.124  & 0.099  \\
            

\midrule
 \multirow{3}*{\shortstack[1]{Pose Network \\ (\textbf{RAUM-VO})}} & $t_{err}$& 2.927 &   5.843 \\
 & ATE &  8.664 &  12.297 \\
 & RPE (m) &   0.068 &   0.078 \\
\bottomrule
\end{tabular}
}
\end{center}
\vspace{-10pt}
\label{table:scale_align}
\end{table}

%% file: latex/content/conclusion.tex
\section{Conclusion}
\label{sec:con}
In this paper, we have presented our approach, RAUM-VO, that combines the translation predicted by a Pose Network with the rotations estimated by a geometrical method named F2F. In practice, we introduced an additional self-supervised loss to guide the training. More importantly, during online inference, we adjust the rotations predicted by the Pose Network with a single estimation of F2F, avoiding complex strategies for model selection and multiple RANSAC loops. Finally, we evaluated RAUM-VO on the KITTI odometry dataset and compared it with other relevant state-of-the-art methods. While efficient, this adjustment step is decisive for improving the prediction of unsupervised pose networks.

Future works can track or match these features, using the associated descriptor, over longer frame distances enabling local or global-BA with loop closures similar to ORB-SLAM. More interestingly, the extension of F2F to multiple views proposed by Lee and Civera~\cite{RotationBundleAdj} could be an alternative to Rotation Averaging~\cite{HartleyRotavg} to initialize the pose graph optimization~\cite{Initializationfor3dSLAM} together with the Pose Network prediction.

%% file: main.bbl
\begin{thebibliography}{10}

\bibitem{galvez2012bags}
Dorian G{\'a}lvez-L{\'o}pez and Juan~D Tardos.
\newblock Bags of binary words for fast place recognition in image sequences.
\newblock {\em IEEE Transactions on Robotics}, 28(5):1188--1197, 2012.

\bibitem{dellaert2017factor}
Frank Dellaert and Michael Kaess.
\newblock Factor graphs for robot perception.
\newblock {\em Found. Trends Robotics}, 6(1-2):1--139, 2017.

\bibitem{VOTuto}
Davide Scaramuzza and Friedrich Fraundorfer.
\newblock Visual odometry [tutorial].
\newblock {\em {IEEE} Robotics Autom. Mag.}, 18(4):80--92, 2011.

\bibitem{SLAMSurvey}
Takafumi Taketomi, Hideaki Uchiyama, and Sei Ikeda.
\newblock Visual {SLAM} algorithms: a survey from 2010 to 2016.
\newblock {\em {IPSJ} Trans. Comput. Vis. Appl.}, 9:16, 2017.

\bibitem{PTAM}
Georg Klein and David~William Murray.
\newblock Parallel tracking and mapping for small {AR} workspaces.
\newblock In {\em Sixth {IEEE/ACM} International Symposium on Mixed and
  Augmented Reality, {ISMAR} 2007, 13-16 November 2007, Nara, Japan}, pages
  225--234. {IEEE} Computer Society, 2007.

\bibitem{VideoBasedSLAM}
George Vogiatzis and Carlos Hern{\'{a}}ndez.
\newblock Video-based, real-time multi-view stereo.
\newblock {\em Image Vis. Comput.}, 29(7):434--441, 2011.

\bibitem{SemiDenseVO}
Jakob Engel, J{\"{u}}rgen Sturm, and Daniel Cremers.
\newblock Semi-dense visual odometry for a monocular camera.
\newblock In {\em {IEEE} International Conference on Computer Vision, {ICCV}
  2013, Sydney, Australia, December 1-8, 2013}, pages 1449--1456. {IEEE}
  Computer Society, 2013.

\bibitem{DeepVOSurvey}
Yue Ming, Xuyang Meng, Chunxiao Fan, and Hui Yu.
\newblock Deep learning for monocular depth estimation: {A} review.
\newblock {\em Neurocomputing}, 438:14--33, 2021.

\bibitem{TwoStreamNet}
Rares Ambrus, Vitor Guizilini, Jie Li, Sudeep Pillai, and Adrien Gaidon.
\newblock Two stream networks for self-supervised ego-motion estimation.
\newblock In Leslie~Pack Kaelbling, Danica Kragic, and Komei Sugiura, editors,
  {\em 3rd Annual Conference on Robot Learning, CoRL 2019, Osaka, Japan,
  October 30 - November 1, 2019, Proceedings}, volume 100 of {\em Proceedings
  of Machine Learning Research}, pages 1052--1061. {PMLR}, 2019.

\bibitem{VISO2}
Andreas Geiger, Julius Ziegler, and Christoph Stiller.
\newblock Stereoscan: Dense 3d reconstruction in real-time.
\newblock In {\em {IEEE} Intelligent Vehicles Symposium (IV), 2011,
  Baden-Baden, Germany, June 5-9, 2011}, pages 963--968. {IEEE}, 2011.

\bibitem{ORB_SLAM}
Raul Mur{-}Artal, J.~M.~M. Montiel, and Juan~D. Tard{\'{o}}s.
\newblock {ORB-SLAM:} {A} versatile and accurate monocular {SLAM} system.
\newblock {\em {IEEE} Trans. Robotics}, 31(5):1147--1163, 2015.

\bibitem{Zhou17SFMlearner}
Tinghui Zhou, Matthew Brown, Noah Snavely, and David~G. Lowe.
\newblock Unsupervised learning of depth and ego-motion from video.
\newblock In {\em 2017 {IEEE} Conference on Computer Vision and Pattern
  Recognition, {CVPR} 2017, Honolulu, HI, USA, July 21-26, 2017}, pages
  6612--6619. {IEEE} Computer Society, 2017.

\bibitem{DSO}
Jakob Engel, Vladlen Koltun, and Daniel Cremers.
\newblock Direct sparse odometry.
\newblock {\em {IEEE} Trans. Pattern Anal. Mach. Intell.}, 40(3):611--625,
  2018.

\bibitem{MVSBook}
Andrew Harltey and Andrew Zisserman.
\newblock {\em Multiple view geometry in computer vision {(2.} ed.)}.
\newblock Cambridge University Press, 2006.

\bibitem{TowardsGeneralization}
Wang Zhao, Shaohui Liu, Yezhi Shu, and Yong{-}Jin Liu.
\newblock Towards better generalization: Joint depth-pose learning without
  posenet.
\newblock In {\em 2020 {IEEE/CVF} Conference on Computer Vision and Pattern
  Recognition, {CVPR} 2020, Seattle, WA, USA, June 13-19, 2020}, pages
  9148--9158. Computer Vision Foundation / {IEEE}, 2020.

\bibitem{DF-VO}
Huangying Zhan, Chamara~Saroj Weerasekera, Jia{-}Wang Bian, Ravi Garg, and
  Ian~D. Reid.
\newblock {DF-VO:} what should be learnt for visual odometry?
\newblock {\em CoRR}, abs/2103.00933, 2021.

\bibitem{Superpoint}
Daniel DeTone, Tomasz Malisiewicz, and Andrew Rabinovich.
\newblock Superpoint: Self-supervised interest point detection and description.
\newblock In {\em Proceedings of the IEEE conference on computer vision and
  pattern recognition workshops}, pages 224--236, 2018.

\bibitem{Superglue}
Paul-Edouard Sarlin, Daniel DeTone, Tomasz Malisiewicz, and Andrew Rabinovich.
\newblock Superglue: Learning feature matching with graph neural networks.
\newblock In {\em Proceedings of the IEEE/CVF conference on computer vision and
  pattern recognition}, pages 4938--4947, 2020.

\bibitem{KneipTwoViewRotBundleAdj}
Laurent Kneip and Simon Lynen.
\newblock Direct optimization of frame-to-frame rotation.
\newblock In {\em {IEEE} International Conference on Computer Vision, {ICCV}
  2013, Sydney, Australia, December 1-8, 2013}, pages 2352--2359. {IEEE}
  Computer Society, 2013.

\bibitem{Monodepth2}
Cl{\'{e}}ment Godard, Oisin~Mac Aodha, Michael Firman, and Gabriel~J. Brostow.
\newblock Digging into self-supervised monocular depth estimation.
\newblock In {\em 2019 {IEEE/CVF} International Conference on Computer Vision,
  {ICCV} 2019, Seoul, Korea (South), October 27 - November 2, 2019}, pages
  3827--3837. {IEEE}, 2019.

\bibitem{FastSLAM}
Michael Montemerlo, Sebastian Thrun, Daphne Koller, and Ben Wegbreit.
\newblock Fastslam: {A} factored solution to the simultaneous localization and
  mapping problem.
\newblock In Rina Dechter, Michael~J. Kearns, and Richard~S. Sutton, editors,
  {\em Proceedings of the Eighteenth National Conference on Artificial
  Intelligence and Fourteenth Conference on Innovative Applications of
  Artificial Intelligence, July 28 - August 1, 2002, Edmonton, Alberta,
  Canada}, pages 593--598. {AAAI} Press / The {MIT} Press, 2002.

\bibitem{SqrootSAM}
Frank Dellaert and Michael Kaess.
\newblock Square root {SAM:} simultaneous localization and mapping via square
  root information smoothing.
\newblock {\em Int. J. Robotics Res.}, 25(12):1181--1203, 2006.

\bibitem{BundleAdj}
Bill Triggs, Philip~F. McLauchlan, Richard~I. Hartley, and Andrew~W.
  Fitzgibbon.
\newblock Bundle adjustment - {A} modern synthesis.
\newblock In Bill Triggs, Andrew Zisserman, and Richard Szeliski, editors, {\em
  Vision Algorithms: Theory and Practice, International Workshop on Vision
  Algorithms, held during {ICCV} '99, Corfu, Greece, September 21-22, 1999,
  Proceedings}, volume 1883 of {\em Lecture Notes in Computer Science}, pages
  298--372. Springer, 1999.

\bibitem{VIOAR}
Davide Scaramuzza and Zichao Zhang.
\newblock Visual-inertial odometry of aerial robots.
\newblock {\em CoRR}, abs/1906.03289, 2019.

\bibitem{WhyFilter}
Hauke Strasdat, J.~M.~M. Montiel, and Andrew~J. Davison.
\newblock Visual {SLAM:} why filter?
\newblock {\em Image Vis. Comput.}, 30(2):65--77, 2012.

\bibitem{LSD-SLAM}
Jakob Engel, Thomas Sch{\"{o}}ps, and Daniel Cremers.
\newblock {LSD-SLAM:} large-scale direct monocular {SLAM}.
\newblock In David~J. Fleet, Tom{\'{a}}s Pajdla, Bernt Schiele, and Tinne
  Tuytelaars, editors, {\em Computer Vision - {ECCV} 2014 - 13th European
  Conference, Zurich, Switzerland, September 6-12, 2014, Proceedings, Part
  {II}}, volume 8690 of {\em Lecture Notes in Computer Science}, pages
  834--849. Springer, 2014.

\bibitem{Nister5points}
David Nist{\'{e}}r.
\newblock An efficient solution to the five-point relative pose problem.
\newblock {\em {IEEE} Trans. Pattern Anal. Mach. Intell.}, 26(6):756--777,
  2004.

\bibitem{longuet1981computer}
H~Christopher Longuet-Higgins.
\newblock A computer algorithm for reconstructing a scene from two projections.
\newblock {\em Nature}, 293(5828):133--135, 1981.

\bibitem{EPNP}
Vincent Lepetit, Francesc Moreno{-}Noguer, and Pascal Fua.
\newblock Ep\emph{n}p: An accurate \emph{O}(\emph{n}) solution to the
  p\emph{n}p problem.
\newblock {\em Int. J. Comput. Vis.}, 81(2):155--166, 2009.

\bibitem{cantzler1981random}
H~Cantzler.
\newblock Random sample consensus (ransac).
\newblock {\em Institute for Perception, Action and Behaviour, Division of
  Informatics, University of Edinburgh}, 1981.

\bibitem{Garg16}
Ravi Garg, B.~G.~Vijay Kumar, Gustavo Carneiro, and Ian~D. Reid.
\newblock Unsupervised {CNN} for single view depth estimation: Geometry to the
  rescue.
\newblock In Bastian Leibe, Jiri Matas, Nicu Sebe, and Max Welling, editors,
  {\em Computer Vision - {ECCV} 2016 - 14th European Conference, Amsterdam, The
  Netherlands, October 11-14, 2016, Proceedings, Part {VIII}}, volume 9912 of
  {\em Lecture Notes in Computer Science}, pages 740--756. Springer, 2016.

\bibitem{Godard17}
Cl{\'{e}}ment Godard, Oisin~Mac Aodha, and Gabriel~J. Brostow.
\newblock Unsupervised monocular depth estimation with left-right consistency.
\newblock In {\em 2017 {IEEE} Conference on Computer Vision and Pattern
  Recognition, {CVPR} 2017, Honolulu, HI, USA, July 21-26, 2017}, pages
  6602--6611. {IEEE} Computer Society, 2017.

\bibitem{SSIM}
Zhou Wang, Alan~C. Bovik, Hamid~R. Sheikh, and Eero~P. Simoncelli.
\newblock Image quality assessment: from error visibility to structural
  similarity.
\newblock {\em {IEEE} Trans. Image Process.}, 13(4):600--612, 2004.

\bibitem{SpatialTransformerNet}
Max Jaderberg, Karen Simonyan, Andrew Zisserman, and Koray Kavukcuoglu.
\newblock Spatial transformer networks.
\newblock In Corinna Cortes, Neil~D. Lawrence, Daniel~D. Lee, Masashi Sugiyama,
  and Roman Garnett, editors, {\em Advances in Neural Information Processing
  Systems 28: Annual Conference on Neural Information Processing Systems 2015,
  December 7-12, 2015, Montreal, Quebec, Canada}, pages 2017--2025, 2015.

\bibitem{UndeepVO}
Ruihao Li, Sen Wang, Zhiqiang Long, and Dongbing Gu.
\newblock Undeepvo: Monocular visual odometry through unsupervised deep
  learning.
\newblock In {\em 2018 {IEEE} International Conference on Robotics and
  Automation, {ICRA} 2018, Brisbane, Australia, May 21-25, 2018}, pages
  7286--7291. {IEEE}, 2018.

\bibitem{UnsupWFeatRec}
Huangying Zhan, Ravi Garg, Chamara~Saroj Weerasekera, Kejie Li, Harsh Agarwal,
  and Ian~D. Reid.
\newblock Unsupervised learning of monocular depth estimation and visual
  odometry with deep feature reconstruction.
\newblock In {\em 2018 {IEEE} Conference on Computer Vision and Pattern
  Recognition, {CVPR} 2018, Salt Lake City, UT, USA, June 18-22, 2018}, pages
  340--349. Computer Vision Foundation / {IEEE} Computer Society, 2018.

\bibitem{UnsupMono3dICP}
Reza Mahjourian, Martin Wicke, and Anelia Angelova.
\newblock Unsupervised learning of depth and ego-motion from monocular video
  using 3d geometric constraints.
\newblock In {\em 2018 {IEEE} Conference on Computer Vision and Pattern
  Recognition, {CVPR} 2018, Salt Lake City, UT, USA, June 18-22, 2018}, pages
  5667--5675. Computer Vision Foundation / {IEEE} Computer Society, 2018.

\bibitem{Bian19}
Jiawang Bian, Zhichao Li, Naiyan Wang, Huangying Zhan, Chunhua Shen,
  Ming{-}Ming Cheng, and Ian~D. Reid.
\newblock Unsupervised scale-consistent depth and ego-motion learning from
  monocular video.
\newblock In Hanna~M. Wallach, Hugo Larochelle, Alina Beygelzimer, Florence
  d'Alch{\'{e}}{-}Buc, Emily~B. Fox, and Roman Garnett, editors, {\em Advances
  in Neural Information Processing Systems 32: Annual Conference on Neural
  Information Processing Systems 2019, NeurIPS 2019, December 8-14, 2019,
  Vancouver, BC, Canada}, pages 35--45, 2019.

\bibitem{Luo20Consistent}
Xuan Luo, Jia{-}Bin Huang, Richard Szeliski, Kevin Matzen, and Johannes Kopf.
\newblock Consistent video depth estimation.
\newblock {\em {ACM} Trans. Graph.}, 39(4):71, 2020.

\bibitem{GeneralizingDeepVO}
Shunkai Li, Xin Wu, Yingdian Cao, and Hongbin Zha.
\newblock Generalizing to the open world: Deep visual odometry with online
  adaptation.
\newblock In {\em {IEEE} Conference on Computer Vision and Pattern Recognition,
  {CVPR} 2021, virtual, June 19-25, 2021}, pages 13184--13193. Computer Vision
  Foundation / {IEEE}, 2021.

\bibitem{casser2019depth}
Vincent Casser, Soeren Pirk, Reza Mahjourian, and Anelia Angelova.
\newblock Depth prediction without the sensors: Leveraging structure for
  unsupervised learning from monocular videos.
\newblock In {\em Proceedings of the AAAI conference on artificial
  intelligence}, volume~33, pages 8001--8008, 2019.

\bibitem{vijayanarasimhan2017sfm}
Sudheendra Vijayanarasimhan, Susanna Ricco, Cordelia Schmid, Rahul Sukthankar,
  and Katerina Fragkiadaki.
\newblock Sfm-net: Learning of structure and motion from video.
\newblock {\em arXiv preprint arXiv:1704.07804}, 2017.

\bibitem{yin2018geonet}
Zhichao Yin and Jianping Shi.
\newblock Geonet: Unsupervised learning of dense depth, optical flow and camera
  pose.
\newblock In {\em Proceedings of the IEEE conference on computer vision and
  pattern recognition}, pages 1983--1992, 2018.

\bibitem{zou2018dfnet}
Yuliang Zou, Zelun Luo, and Jia-Bin Huang.
\newblock Df-net: Unsupervised joint learning of depth and flow using
  cross-task consistency.
\newblock In {\em Proceedings of the European conference on computer vision
  (ECCV)}, pages 36--53, 2018.

\bibitem{zhao2018learning}
Cheng Zhao, Li~Sun, Pulak Purkait, Tom Duckett, and Rustam Stolkin.
\newblock Learning monocular visual odometry with dense 3d mapping from dense
  3d flow.
\newblock In {\em 2018 IEEE/RSJ International Conference on Intelligent Robots
  and Systems (IROS)}, pages 6864--6871. IEEE, 2018.

\bibitem{lee2019learning}
Seokju Lee, Sunghoon Im, Stephen Lin, and In~So Kweon.
\newblock Learning residual flow as dynamic motion from stereo videos.
\newblock In {\em 2019 IEEE/RSJ International Conference on Intelligent Robots
  and Systems (IROS)}, pages 1180--1186. IEEE, 2019.

\bibitem{ranjan2019competitive}
Anurag Ranjan, Varun Jampani, Lukas Balles, Kihwan Kim, Deqing Sun, Jonas
  Wulff, and Michael~J Black.
\newblock Competitive collaboration: Joint unsupervised learning of depth,
  camera motion, optical flow and motion segmentation.
\newblock In {\em Proceedings of the IEEE/CVF conference on computer vision and
  pattern recognition}, pages 12240--12249, 2019.

\bibitem{EPC++}
Chenxu Luo, Zhenheng Yang, Peng Wang, Yang Wang, Wei Xu, Ram Nevatia, and
  Alan~L. Yuille.
\newblock Every pixel counts ++: Joint learning of geometry and motion with 3d
  holistic understanding.
\newblock {\em {IEEE} Trans. Pattern Anal. Mach. Intell.}, 42(10):2624--2641,
  2020.

\bibitem{chen2019self}
Yuhua Chen, Cordelia Schmid, and Cristian Sminchisescu.
\newblock Self-supervised learning with geometric constraints in monocular
  video: Connecting flow, depth, and camera.
\newblock In {\em Proceedings of the IEEE/CVF International Conference on
  Computer Vision}, pages 7063--7072, 2019.

\bibitem{li2020unsupervised}
Hanhan Li, Ariel Gordon, Hang Zhao, Vincent Casser, and Anelia Angelova.
\newblock Unsupervised monocular depth learning in dynamic scenes.
\newblock {\em arXiv preprint arXiv:2010.16404}, 2020.

\bibitem{wang2021motionhint}
Cong Wang, Yu-Ping Wang, and Dinesh Manocha.
\newblock Motionhint: Self-supervised monocular visual odometry with motion
  constraints.
\newblock {\em arXiv preprint arXiv:2109.06768}, 2021.

\bibitem{jiang2021unsupervised}
Hualie Jiang, Laiyan Ding, Zhenglong Sun, and Rui Huang.
\newblock Unsupervised monocular depth perception: Focusing on moving objects.
\newblock {\em IEEE Sensors Journal}, 21(24):27225--27237, 2021.

\bibitem{SVO}
Christian Forster, Matia Pizzoli, and Davide Scaramuzza.
\newblock {SVO:} fast semi-direct monocular visual odometry.
\newblock In {\em 2014 {IEEE} International Conference on Robotics and
  Automation, {ICRA} 2014, Hong Kong, China, May 31 - June 7, 2014}, pages
  15--22. {IEEE}, 2014.

\bibitem{wang2018learning}
Chaoyang Wang, Jos{\'e}~Miguel Buenaposada, Rui Zhu, and Simon Lucey.
\newblock Learning depth from monocular videos using direct methods.
\newblock In {\em Proceedings of the IEEE Conference on Computer Vision and
  Pattern Recognition}, pages 2022--2030, 2018.

\bibitem{yang2018deep}
Nan Yang, Rui Wang, Jorg Stuckler, and Daniel Cremers.
\newblock Deep virtual stereo odometry: Leveraging deep depth prediction for
  monocular direct sparse odometry.
\newblock In {\em Proceedings of the European Conference on Computer Vision
  (ECCV)}, pages 817--833, 2018.

\bibitem{li2019pose}
Yang Li, Yoshitaka Ushiku, and Tatsuya Harada.
\newblock Pose graph optimization for unsupervised monocular visual odometry.
\newblock In {\em 2019 International Conference on Robotics and Automation
  (ICRA)}, pages 5439--5445. IEEE, 2019.

\bibitem{loo2019cnn}
Shing~Yan Loo, Ali~Jahani Amiri, Syamsiah Mashohor, Sai~Hong Tang, and Hong
  Zhang.
\newblock Cnn-svo: Improving the mapping in semi-direct visual odometry using
  single-image depth prediction.
\newblock In {\em 2019 International Conference on Robotics and Automation
  (ICRA)}, pages 5218--5223. IEEE, 2019.

\bibitem{tiwari2020pseudo}
Lokender Tiwari, Pan Ji, Quoc-Huy Tran, Bingbing Zhuang, Saket Anand, and
  Manmohan Chandraker.
\newblock Pseudo rgb-d for self-improving monocular slam and depth prediction.
\newblock In {\em European Conference on Computer Vision}, pages 437--455.
  Springer, 2020.

\bibitem{cheng2020depth}
Ran Cheng, Christopher Agia, David Meger, and Gregory Dudek.
\newblock Depth prediction for monocular direct visual odometry.
\newblock In {\em 2020 17th Conference on Computer and Robot Vision (CRV)},
  pages 70--77. IEEE Computer Society, 2020.

\bibitem{bian2021unsupervised}
Jia-Wang Bian, Huangying Zhan, Naiyan Wang, Zhichao Li, Le~Zhang, Chunhua Shen,
  Ming-Ming Cheng, and Ian Reid.
\newblock Unsupervised scale-consistent depth learning from video.
\newblock {\em International Journal of Computer Vision}, 129(9):2548--2564,
  2021.

\bibitem{D3VO}
Nan Yang, Lukas von Stumberg, Rui Wang, and Daniel Cremers.
\newblock {D3VO:} deep depth, deep pose and deep uncertainty for monocular
  visual odometry.
\newblock In {\em 2020 {IEEE/CVF} Conference on Computer Vision and Pattern
  Recognition, {CVPR} 2020, Seattle, WA, USA, June 13-19, 2020}, pages
  1278--1289. Computer Vision Foundation / {IEEE}, 2020.

\bibitem{visualloss}
H~Zhao, O~Gallo, I~Frosio, and J~Kautz.
\newblock Is l2 a good loss function for neural networks for image processing?
  arxiv preprint.
\newblock {\em arXiv preprint arXiv:1511.08861}, 2015.

\bibitem{strasdat2010scale}
Hauke Strasdat, J~Montiel, and Andrew~J Davison.
\newblock Scale drift-aware large scale monocular slam.
\newblock {\em Robotics: Science and Systems VI}, 2(3):7, 2010.

\bibitem{CNN-SLAM}
Keisuke Tateno, Federico Tombari, Iro Laina, and Nassir Navab.
\newblock {CNN-SLAM:} real-time dense monocular {SLAM} with learned depth
  prediction.
\newblock In {\em 2017 {IEEE} Conference on Computer Vision and Pattern
  Recognition, {CVPR} 2017, Honolulu, HI, USA, July 21-26, 2017}, pages
  6565--6574. {IEEE} Computer Society, 2017.

\bibitem{3dpacking}
Vitor Guizilini, Rares Ambrus, Sudeep Pillai, Allan Raventos, and Adrien
  Gaidon.
\newblock 3d packing for self-supervised monocular depth estimation.
\newblock In {\em 2020 {IEEE/CVF} Conference on Computer Vision and Pattern
  Recognition, {CVPR} 2020, Seattle, WA, USA, June 13-19, 2020}, pages
  2482--2491. Computer Vision Foundation / {IEEE}, 2020.

\bibitem{TwoViewRotation}
Laurent Kneip, Roland Siegwart, and Marc Pollefeys.
\newblock Finding the exact rotation between two images independently of the
  translation.
\newblock In Andrew~W. Fitzgibbon, Svetlana Lazebnik, Pietro Perona, Yoichi
  Sato, and Cordelia Schmid, editors, {\em Computer Vision - {ECCV} 2012 - 12th
  European Conference on Computer Vision, Florence, Italy, October 7-13, 2012,
  Proceedings, Part {VI}}, volume 7577 of {\em Lecture Notes in Computer
  Science}, pages 696--709. Springer, 2012.

\bibitem{gao2021introduction}
Xiang Gao and Tao Zhang.
\newblock {\em Introduction to Visual SLAM: From Theory to Practice}.
\newblock Springer Nature, 2021.

\bibitem{kneip2014opengv}
Laurent Kneip and Paul Furgale.
\newblock Opengv: A unified and generalized approach to real-time calibrated
  geometric vision.
\newblock In {\em 2014 IEEE International Conference on Robotics and Automation
  (ICRA)}, pages 1--8. IEEE, 2014.

\bibitem{UNET}
Olaf Ronneberger, Philipp Fischer, and Thomas Brox.
\newblock U-net: Convolutional networks for biomedical image segmentation.
\newblock In Nassir Navab, Joachim Hornegger, William M.~Wells III, and
  Alejandro~F. Frangi, editors, {\em Medical Image Computing and
  Computer-Assisted Intervention - {MICCAI} 2015 - 18th International
  Conference Munich, Germany, October 5 - 9, 2015, Proceedings, Part {III}},
  volume 9351 of {\em Lecture Notes in Computer Science}, pages 234--241.
  Springer, 2015.

\bibitem{mayer2016large}
Nikolaus Mayer, Eddy Ilg, Philip Hausser, Philipp Fischer, Daniel Cremers,
  Alexey Dosovitskiy, and Thomas Brox.
\newblock A large dataset to train convolutional networks for disparity,
  optical flow, and scene flow estimation.
\newblock In {\em Proceedings of the IEEE conference on computer vision and
  pattern recognition}, pages 4040--4048, 2016.

\bibitem{resnet}
Kaiming He, Xiangyu Zhang, Shaoqing Ren, and Jian Sun.
\newblock Deep residual learning for image recognition.
\newblock In {\em Proceedings of the IEEE conference on computer vision and
  pattern recognition}, pages 770--778, 2016.

\bibitem{elu}
Djork-Arn{\'e} Clevert, Thomas Unterthiner, and Sepp Hochreiter.
\newblock Fast and accurate deep network learning by exponential linear units
  (elus).
\newblock {\em arXiv preprint arXiv:1511.07289}, 2015.

\bibitem{relu}
Abien~Fred Agarap.
\newblock Deep learning using rectified linear units (relu).
\newblock {\em arXiv preprint arXiv:1803.08375}, 2018.

\bibitem{paszke2019pytorch}
Adam Paszke, Sam Gross, Francisco Massa, Adam Lerer, James Bradbury, Gregory
  Chanan, Trevor Killeen, Zeming Lin, Natalia Gimelshein, Luca Antiga, et~al.
\newblock Pytorch: An imperative style, high-performance deep learning library.
\newblock {\em Advances in neural information processing systems}, 32, 2019.

\bibitem{deng2009imagenet}
Jia Deng, Wei Dong, Richard Socher, Li-Jia Li, Kai Li, and Li~Fei-Fei.
\newblock Imagenet: A large-scale hierarchical image database.
\newblock In {\em 2009 IEEE conference on computer vision and pattern
  recognition}, pages 248--255. Ieee, 2009.

\bibitem{kingma2014adam}
Diederik~P Kingma and Jimmy Ba.
\newblock Adam: A method for stochastic optimization.
\newblock {\em arXiv preprint arXiv:1412.6980}, 2014.

\bibitem{KITTI}
Andreas Geiger, Philip Lenz, Christoph Stiller, and Raquel Urtasun.
\newblock Vision meets robotics: The kitti dataset.
\newblock {\em The International Journal of Robotics Research},
  32(11):1231--1237, 2013.

\bibitem{umeyama1991least}
Shinji Umeyama.
\newblock Least-squares estimation of transformation parameters between two
  point patterns.
\newblock {\em IEEE Transactions on Pattern Analysis \& Machine Intelligence},
  13(04):376--380, 1991.

\bibitem{lee2013smooth}
John~M Lee.
\newblock Smooth manifolds.
\newblock In {\em Introduction to Smooth Manifolds}, pages 1--31. Springer,
  2013.

\bibitem{zhou2019continuity}
Yi~Zhou, Connelly Barnes, Jingwan Lu, Jimei Yang, and Hao Li.
\newblock On the continuity of rotation representations in neural networks.
\newblock In {\em Proceedings of the IEEE/CVF Conference on Computer Vision and
  Pattern Recognition}, pages 5745--5753, 2019.

\bibitem{huynh2009metrics}
Du~Q Huynh.
\newblock Metrics for 3d rotations: Comparison and analysis.
\newblock {\em Journal of Mathematical Imaging and Vision}, 35(2):155--164,
  2009.

\bibitem{shen2019beyond}
Tianwei Shen, Zixin Luo, Lei Zhou, Hanyu Deng, Runze Zhang, Tian Fang, and Long
  Quan.
\newblock Beyond photometric loss for self-supervised ego-motion estimation.
\newblock In {\em 2019 International Conference on Robotics and Automation
  (ICRA)}, pages 6359--6365. IEEE, 2019.

\bibitem{RotationBundleAdj}
Seong~Hun Lee and Javier Civera.
\newblock Rotation-only bundle adjustment.
\newblock In {\em {IEEE} Conference on Computer Vision and Pattern Recognition,
  {CVPR} 2021, virtual, June 19-25, 2021}, pages 424--433. Computer Vision
  Foundation / {IEEE}, 2021.

\bibitem{HartleyRotavg}
Richard~I. Hartley, Khurrum Aftab, and Jochen Trumpf.
\newblock {L1} rotation averaging using the weiszfeld algorithm.
\newblock In {\em The 24th {IEEE} Conference on Computer Vision and Pattern
  Recognition, {CVPR} 2011, Colorado Springs, CO, USA, 20-25 June 2011}, pages
  3041--3048. {IEEE} Computer Society, 2011.

\bibitem{Initializationfor3dSLAM}
Luca Carlone, Roberto Tron, Kostas Daniilidis, and Frank Dellaert.
\newblock Initialization techniques for 3d {SLAM:} {A} survey on rotation
  estimation and its use in pose graph optimization.
\newblock In {\em {IEEE} International Conference on Robotics and Automation,
  {ICRA} 2015, Seattle, WA, USA, 26-30 May, 2015}, pages 4597--4604. {IEEE},
  2015.

\end{thebibliography}
